\RequirePackage{fix-cm}
\documentclass[smallcondensed]{svjour3} % onecolumn (ditto)

\usepackage[round]{natbib}

\usepackage{amsmath}
\usepackage[mathscr]{euscript}

\usepackage{hyphenat}
\usepackage{booktabs}
\usepackage{algorithm}
\usepackage{algpseudocode}

\usepackage{amsmath}
\usepackage{amssymb}

\usepackage{graphicx}
\usepackage{subfiles}

\usepackage{subfig}

\usepackage{tabularx}

\usepackage[bookmarks=true,bookmarksdepth=3]{hyperref}

\usepackage[shortcuts]{extdash}

\newtheorem{exp-hypothesis}{Experimental Hypothesis}
\newtheorem{observation}{Observation}
\DeclareMathSymbol{\mlq}{\mathord}{operators}{``}
\DeclareMathSymbol{\mrq}{\mathord}{operators}{`'}
\begin{document}

\title{Meta-Interpretive Learning as Metarule Specialisation}

\titlerunning{Meta-Interpretive Learning as Metarule Specialisation}

\author{ S. Patsantzis \and 
	 S.H. Muggleton
}

\institute{S. Patsantzis \at
	   Imperial College London \\
	   United Kingdom \\
	   \email{ep2216@ic.ac.uk
	}
	\and
	   S. H. Muggleton \at
	   Imperial College London \\
	   United Kingdom \\
	   \email{s.muggleton@imperial.ac.uk
	}
}

\date{Received: / Accepted: date}

\maketitle

\begin{abstract}
In Meta-Interpretive Learning (MIL) the metarules, second-order datalog clauses
acting as inductive bias, are manually defined by the user. In this work we show
that second-order metarules for MIL can be learned by MIL. We define a
generality ordering of metarules by $\theta$-subsumption and show that
user-defined \emph{sort metarules} are derivable by specialisation of the
most-general \emph{matrix metarules} in a language class; and that these matrix
metarules are in turn derivable by specialisation of third-order \emph{punch
metarules} with variables quantified over the set of atoms and for which only an
upper bound on their number of literals need be user-defined.  We show that the
cardinality of a metarule language is polynomial in the number of literals in
punch metarules. We re-frame MIL as metarule specialisation by resolution. We
modify the MIL metarule specialisation operator to return new metarules rather
than first-order clauses and prove the correctness of the new operator. We
implement the new operator as TOIL, a sub-system of the MIL system Louise.  Our
experiments show that as user-defined sort metarules are progressively replaced
by sort metarules learned by TOIL, Louise's predictive accuracy and training
times are maintained. We conclude that automatically derived metarules can
replace user-defined metarules.
\end{abstract}

\section{Introduction}
\label{sec:Introduction}

\begin{table}[t]
	\centering
	\begin{tabularx}{\columnwidth}{lllll}
		\multicolumn{2}{l}{\textbf{3rd-order}} & \multicolumn{2}{l}{\textbf{2nd-order}} & \textbf{2nd-order} \\
		\multicolumn{2}{l}{\textbf{Punch}} & \multicolumn{2}{l}{\textbf{Matrix metarules}} & \textbf{Sort metarules} \\
		\multicolumn{2}{l}{\textbf{metarules}} & \multicolumn{2}{l}{\textbf{(most-general)}} & \textbf{(fully-connected)} \\
		\toprule
		$P \leftarrow Q$ & $\preceq$ & $P(x,y) \leftarrow Q(z,u)$ & $\preceq$ & $P(x,y) \leftarrow Q(x,y)$ \\
		                 & &                                      & & $P(x,y) \leftarrow Q(y,x)$ \\
		                 & &                                      & & $P(x,y) \leftarrow Q(x,z), R(z,y)$ \\
		                 & &                                      & & ...\\
		$P \leftarrow Q, R$ & $\preceq$ & $P(x,y,z) \leftarrow Q(u,v), R(w,s)$ & $\preceq$ & $P(x,y,z) \leftarrow Q(x,z), R(z,y)$ \\
				 & &                                      & & $P(x,y,z) \leftarrow Q(x,y), R(y,z)$ \\
				 & &                                      & & $P(x,y,z) \leftarrow Q(x,y), R(y,v), S(v,z)$ \\
				 & &                                      & & ....\\
		\bottomrule
	\end{tabularx}
\caption{
Metarule ordering by $\theta$-subsumption. Left column: 3rd-order ``punch
metarules" with variables $P,Q,R$ existentially quantified over the set of
atoms. Middle column: most-general 2nd-order ``matrix metarules" with
variables $P,Q,R$ existentially quantified over predicate symbols and variables
$x,y,z,u,v,w,s$ universally quantified over  constants. Right column:
fully-connected 2nd-order ``sort metarules" with variables $P,Q,R$ existentially
quantified over predicate symbols and $x,y,z,v$ universally quantified over
constants. Variables with the same name in different metarules are distinct
variables. Note subsumption between punch metarules and metarules with arbitrary
numbers of literals of arbitrary arities.}
\label{tab:metarule_ordering}
\end{table}

Meta-Interpretive Learning (MIL) \citep{Muggleton2014,Muggleton2015} is a recent
approach to Inductive Logic Programming (ILP) \citep{Muggleton1994} capable of
learning logic programs with recursive clauses and invented predicates from
examples, background knowledge and a declarative bias, the ``metarules".
Metarules are datalog clauses with variables quantified over predicate symbols
that are therefore \emph{second-order} clauses.

Metarules are often interpreted as language bias or ``clause templates" in
earlier work, particularly outside the MIL literature, but they are in truth
second-order background knowledge used in mixed-order resolution with the
first-order background knowledge to derive the first-order clauses of a
hypothesis that explains the training examples. Resolution with second-order
metarules is made decidable thanks to their encapsulation into first-order
definite clauses, e.g. a metarule $P(x,y) \leftarrow Q(x,y)$ is encapsulated as
a definite clause $m(P,x,y) \leftarrow m(Q,x,y)$.

In the MIL literature, metarules are typically defined by a user according to
intuition or knowledge of a problem domain. A common criticism of MIL is the
dearth of principled approaches for metarule selection (see e.g.
\cite{Cropper2018}). In this work, we formalise MIL as metarule specialisation
by resolution and thereby provide a principled approach to \emph{learning} new
metarules from examples, background knowledge and maximally general metarules.
% Maybe "unique maximally general metarules"?

In particular, we show that user-defined, fully-connected \emph{sort metarules}
commonly used in the MIL literature can be derived automatically by
specialisation of maximally general second-order \emph{matrix metarules}; and
that matrix metarules can be themselves derived by specialisation of
third-order \emph{punch metarules}. This specialisation can be performed by a
standard MIL clause construction operator modified to return metarules rather
than first-order clauses. Predicate invention, performed in MIL by resolution
between metarules, is equivalent to a derivation of new metarules with
arbitrary numbers of literals. Additionally, specialisation of third-order,
punch metarules imposes no restriction on the arities of literals of the
derived metarules. These two capabilities combined finally liberate MIL from
the confines of the $H^2_2$ language fragment of metarules with up to two body
literals of arity at most 2, that is almost exclusively used in the literature
and that is expressive but difficult to use in practice. Table
\ref{tab:metarule_ordering} illustrates punch, matrix and sort metarules and
their generality relations, highlighting the subsumption ordering of
second-order metarules with arbitrary numbers of literals of arbitrary arities.
We illustrate learning metarules outside $H^2_2$ with a worked example in
Section \ref{subsec:Future work}. We present an example of learning in $H^2_2$
in Appendix B.

While the number of specialisations of third order punch metarules can grow very
large, they can be derived efficiently by Top Program Construction (TPC)
\citep{Patsantzis2021a}, a polynomial - time MIL algorithm that forms the basis
of the MIL system Louise \citep{Louise}. We implement metarule learning by TPC
in Louise as a new sub-system called TOIL.\\

We make the following contributions:

\begin{itemize}
	\item We define a generality ordering of third- and second-order metarules
		and first-order clauses, by $\theta$-subsumption.
	\item We prove that metarules and first-order clauses are derivable by
		specialisation of more general metarules.
	\item We redefine MIL as metarule specialisation by SLD-resolution.
	\item We propose a modified MIL specialisation operator to return
		metarules rather than first-order clauses and prove its
		correctness.
	\item We prove that sets of metarule specialisations are polynomial-time enumerable.
	\item We implement our modified operator as a sub-system of Louise
		called TOIL.
	\item We verify experimentally that when user-defined metarules are
		replaced by metarules learned by TOIL, Louise maintains its
		predictive accuracy at the cost of a small increase in training
		times.
\end{itemize}

In Section \ref{sec:Related work} we discuss relevant earlier work. In Section
\ref{sec:Background} we give some background on MIL. In Section
\ref{sec:Framework} we develop the framework of MIL as metarule specialisation
and derive our main theoretical results. In Section \ref{sec:Implementation} we
describe TOIL. In Section \ref{sec:Experiments} we compare Louise's performance
with user-defined and TOIL-learned metarules. We summarise our findings in
Section \ref{sec:Conclusions and future work} and propose future work.

\section{Related work}
\label{sec:Related work}

% TODO: CLINT may need a reference: it can extract metarules from previously
% learned clauses and continue learning.

% TODO: This is the first attempt to present a grand, unified theory of MIL as
% metarule specialisation, that also places it in the standard context of ILP
% algorithm analysis as bottom-up or top-down (from the specific to the general
% or from the general to the specific), in terms of generalisation. The fact of
% that should be pointed out here and in the contributions section, but it
% should also be made clear in the Background section.

% TODO: Interesting, but probably best in Future Work section.
%\citep{Kietz92} interestingly note that the generalisation hierarchy of
%metarules is independent of the learning task and can be precomputed to save
%future computing resources. Unfortunately, precomputing a large number of
%metarules for use with any learning problem for which most are likely to be
%irrelevant can result in over-generalisation and a severe increase of the
%complexity of learning. Our proposed approach of metarule specialisation by MIL
%learns new metarules that are specific to a learning task thus avoiding the
%costs of irrelevance, but also leaves open the prospect of learning metarules
%appropriate to \emph{classes} of learning tasks, in other words a process of
%inductive bias discovery.

Declarative bias in clausal form, similar to metarules, is common in machine
learing. \cite{Emde83} propose the use of Horn clauses to declare transitivity,
conversity, or parallelism relations between binary predicates, however these
early ``metarules" are first-order definite clauses with ground predicate
symbols. A related approach of ``rule models", having variables in place of
predicate symbols, is developed in later systems METAXA.3 \citep{Emde87}, BLIP
\citep{Wrobel1988} and MOBAL \citep{Morik1993,Kietz92}. More recent work in ILP
and program synthesis uses metarules as templates to restrict the hypothesis
search space, for example the work by \cite{Evans2018} or \cite{Xujie2018}, and
\cite{Xujie2019} which draw inspiration from MIL but use metarules only as
templates.

By contrast, metarules are used in MIL not as ``templates" with ``blanks" to be
``filled in", but as second-order formulae with the structure of datalog clauses
that are resolved with first-order background knowledge to refute examples and
derive the first-order clauses of a hypothesis. This use of metarules in
reasoning is unique to MIL, which includes our present work.

Our work improves on the earlier use of metarules in MIL and ILP in several
ways. Firstly, we extend the concept of metarules to include third-order
metarules with variables quantified over the set of atoms. Such metarules are
too general to be used efficiently as clause templates and are best understood
within a generality framework that relates them to second-order metarules.
Secondly, we define such a generality ordering by $\theta$-subsumption over
third- and second-order metarules and first-order definite clauses.
\cite{Kietz92} define a generality ordering between ``rule models" as a special
case of $\theta$-subsumption \citep{Plotkin1972} restricted to variable
instantiation. Our framework instead extends the full definition of
$\theta-$subsumption to metarules. \cite{Xujie2018} also extend
$\theta$-subsumption to metarules without restriction, but maintain two
disjoint generality orders: one for metarules; and one for first-order clauses.
Our work is the first to consider third-order metarules and place third- and
second-order metarules and first-order clauses in a single ordering. Thirdly, we
formalise the description of MIL as metarule specialisation by resolution and
thus explain clause construction and predicate invention in MIL as the
application of a metarule specialisation operator.

Our main contribution is the modification of the MIL specialisation operator to
derive new metarules by specialisation of more general metarules, thus learning
second-order theories. Previous work on MIL requires metarules to be defined
manually, by intuition or domain knowledge, but our approach learns metarules
from examples and first-order background knowledge. Automatic selection of
metarules for MIL by logical reduction of a metarule language is studied by
\cite{Cropper2015} and \cite{Cropper2018}. Unlike this earlier work, our
approach learns new metarules rather than selecting from a user-defined set and
does not suffer from the irreducibility of metarule languages to minimal sets
reported by \cite{Cropper2018}. Our approach naturally reduces each metarule
language to a minimal set of most-general metarules from which all other
metarules in the same language can be derived. Further, in our approach, the
number of literals in third-order metarules suffices to define a metarule
language. Finally, previous work on MIL has remained restricted to the $H^2_2$
language, of datalog metarules with at most two body literals of arity at most
2, which is expressive but difficult to use in practice\footnote{For example it
is natural to define addition by an arity-3 predicate $sum(x,y,z)$. The same
can be expressed as an arity-2 predicate $sum([x,y], z)$ but this is not
datalog. The function symbol $[]$ can be removed by flattening
\citep{Rouveirol1994}, requiring two new body literals and two new background
predicates, e.g. $sum(XY,z) \leftarrow head(XY,x), tail(XY,y), ...$ but this is
not in $H^2_2$. We leave an $H^2_2$ datalog definition of $sum$ as an exercise
to the reader.}. Our approach is capable of learning metarules without
restriction to the numbers of literals or their arities.

Our work is comparable to recent work by \cite{Cropper2021}, on the system
Popper, and \cite{Xujie2019} on the system ALPS, both of which are capable of
learning inductive bias, the latter in the form of metarules. Both of those
systems employ generate-and-test approaches guided by $\theta$-subsumption. Our
approach differs to Popper and ALPS in that it is an application of SLD
resolution to second- and third-order clauses, rather than a generate-and-test
approach.

Our implementation extends the Top Program Construction algorithm
\citep{Patsantzis2021a} that avoids an expensive search of a potentially large
hypothesis space and instead constructs all clauses that entail an example with
respect to background knowledge. We extend this earlier work with the ability to
construct second-order clauses without compromising efficiency.

\section{Background}
\label{sec:Background}

% TODO: name definitions, lemmas and theorems.

\subsection{Logical notation}
\label{subsec:Logical notation}

In this section, we extend the terminology established in
\cite{Nienhuys-Cheng1997} with MIL-specific terminology for second- and
third-order definite clauses and programs. We only describe the salient terms in
the nomenclature.

We define a language of first- and higher-order logic programs composed of
clauses, themselves composed of terms, as follows. $\mathcal{P,F,C,A}$ are
disjoint sets of predicate symbols, function symbols, constants and atoms,
respectively. Variables are quantified over (the elements of) a set. A variable
is a term. A constant is a term. If $F$ is a function symbol or a variable
quantified over $\mathcal{F}$, and $t_1, ..., t_n$ are terms, then $F(t_1, ...,
t_n)$ is a term, and $n$ is the arity of $F$. If $P$ is a predicate symbol or a
variable quantified over $\mathcal{P}$, and $t_1, ..., t_n$ are terms, then
$P(t_1, ..., t_n)$ is an atomic formula, or simply atom, and $n$ is the arity of
$P$. Arities are natural numbers. Comma-separated terms in parentheses are the
arguments of a term or an atom. Terms with $n > 0$ arguments are functional
terms, or simply functions. Constants are functions with 0 arguments.

A literal is an atom, or the negation of an atom, or a variable quantified over
$\mathcal{A}$, or the negation of a variable quantified over $\mathcal{A}$. A
clause is a set of literals, interpreted as a disjunction. A logic program, or
simply program, is a set of clauses, interpreted as a conjunction. A clause is
Horn if it has at most one positive literal. A Horn clause is definite if it has
exactly one positive literal, otherwise it is a Horn goal. A literal is datalog
\citep{Ceri1989} if it contains no functions other than constants. A Horn clause
is datalog if it contains only datalog literals and each variable in a positive
literal is shared with at least one negative literal. A logic program is
definite if it contains only definite clauses and datalog if it contains only
datalog clauses.

Terms without variables are ground. Atoms with only ground terms are ground.
Clauses with only ground atoms are ground. Ground terms, atoms and clauses are
0'th-order. Variables quantified over $\mathcal{C}$ are first-order, variables
quantified over $\mathcal{P}$ or $\mathcal{F}$ are second-order, and variables
quantified over $\mathcal{A}$ are third-order. Functions and atoms with order
$k$ arguments are order $k$. Clauses with order $k$ literals are order $k$.
Programs with order $k$ clauses are order $k$. A non-ground order $k$ term, atom
or clause, or an order $k$ program is also order $k-1$.

We denote a variable $X$ quantified over a set $S$ as $\exists_{\in S} X$ or
$\forall_{\in S} X$.

A substitution of variables in a clause $C$ is a finite set $\vartheta =
\{x_1/t_1, ..., x_n/t_n\}$ mapping each variable $x_i$ in $C$ to a term, atom,
or symbol $t_i$. $C\vartheta$, the application of $\vartheta$ to $C$, replaces
each occurrence of $x_i$ with $t_i$ simultaneously.

In keeping with logic programming convention, we will write clauses as
implications, e.g. the definite clause ${A \vee \neg B \vee \neg C}$ will be
written as $A \leftarrow B, C$ where the comma, ``,", indicates a conjunction;
and refer to the positive and negative literals in a clause as the ``head" and
``body" of the clause, respectively.

\subsection{Meta-Interpretive Learning}
\label{subsec:Meta-Interpretive Learning}

MIL is an approach to ILP where programs are learned by specialisation of a set
of higher-order \emph{metarules}. Metarules are defined in the MIL literature as
second-order definite datalog clauses with existentially quantified variables in
place of predicate symbols and constants \citep{Muggleton2015}. In Section
\ref{sec:Framework} we introduce third-order metarules, with variables
quantified over $\mathcal{A}$ as literals.

A system that performs MIL is a Meta-Interpretive Learner, or MIL-learner (with
a slight abuse of abbreviation to allow a natural pronunciation). Examples of
MIL systems are Metagol \citep{metagol}, Thelma \citep{Thelma}, Louise
\citep{Louise} and Hexmil \citep{Kaminski2018ExploitingAS}. A MIL-learner is
given the elements of a MIL problem and returns a \emph{hypothesis}, a logic
program constructed as a solution to the MIL problem. A MIL problem is a
sextuple, $\mathcal{T} = \langle E^+, E^-, B, \mathcal{M}, I, \mathcal{H}
\rangle$ where: a) $E^+$ is a set of ground atoms and $E^-$ is a set of negated
ground atoms of one or more \emph{target predicates}, the positive and negative
examples, respectively; b) $B$ is the background knowledge, a set of definite
clause definitions with datalog heads; c) $\mathcal{M}$ is a set of second-order
metarules; d) $I$ is a set of additional symbols reserved for invented
predicates not defined in $B$ or $E^+$; and e) $\mathcal{H}$ is the hypothesis
space, a set of hypotheses. 

Each hypothesis $H$ in $\mathcal{H}$ is a set of datalog clauses, a definition
of one or more target predicates in $E^+$, and may include definitions of one or
more predicates in $I$. For each $H \in \mathcal{H}$, if $H \wedge B \models
E^+$ and $\forall e^- \in E^-: H \wedge B \not \models e^-$, then $H$ is a
correct hypothesis.

The set, $\mathcal{L}$, of clauses in all hypotheses in $\mathcal{H}$ is the
\emph{Hypothesis Language}. For each clause $C \in \mathcal{L}$, there exists a
metarule $M \in \mathcal{M}$ such that $M \wedge B \wedge E^+ \models C$, i.e.
each clause in $\mathcal{L}$ is an instance of a metarule in $\mathcal{M}$ with
second-order existentially quantified variables substituted for symbols in
$\mathcal{P}$ and first-order existentially quantified variables substituted for
constants in $\mathcal{C}$. $\mathcal{P}$ and $\mathcal{C}$ are populated from
the symbols and constants in $B, E^+$ and $I$.

Typically a MIL learner is not explicitly given $\mathcal{H}$ or $\mathcal{L}$,
rather those are implicitly defined by $\mathcal{M}$ and the constants
$\mathcal{C}$ and symbols $\mathcal{P}$. 

A substitution of the existentially quantified variables in a metarule $M$ is a
\emph{metasubstitution} of $M$. By logic programming convention, a substitution
of universally quantified variables is denoted by a lower-case letter,
$\vartheta, \sigma, \omega$. We follow the convention and also denote
metasubstitutions with \emph{capital} Greek letters, $\Theta, \Sigma, \Omega$,
etc. $\vartheta \Theta$ denotes the composition of the substitution $\vartheta$
and metasubstitution $\Theta$. For brevity, we refer to the composition of a
substitution and metasubstitution as meta/substitution. To help the reader
distinguish a meta/substitution from its application to a metarule we note a
meta/substitution as, e.g. $\vartheta \Theta/M$, whereas $M\vartheta \Theta$ is
the result of applying $\vartheta \Theta$ to $M$.

% TODO: this is a great place to reference earlier correctness results on MIL,
% which we may need in proofs in subsequent sections.
MIL learners construct clauses in $H$ during refutation-proof of a set of
positive examples by SLD-resolution \citep{Nienhuys-Cheng1997} with $B$ and
$\mathcal{M}$. Resolution is performed by a meta-interpreter designed to
preserve the metasubstitutions of metarules in a successful proof by refutation,
while discarding the substitutions of universally quantified variables to avoid
over-specialisation. Metasubstitutions applied to their corresponding metarules
are first-order definite clauses. In a ``second pass" negative examples are
refuted by the same meta-interpreter with $B$, $H$ and $\mathcal{M}$, and any
clauses in $H$ found to entail a negative example are either removed from $H$
(Louise) or replaced on backtracking (Metagol).

Resolution between second-order metarules and first-order clauses is made
decidable by \emph{encapsulation}, a mapping from metarules with existentially
and universally quantified second- and first-order variables, to definite
clauses with only universally quantified, first order variables. For example,
the metarule $\exists P,Q \forall x,y: P(x,y) \leftarrow Q(x,y)$ is encapsulated
as the first-order definite clause $\forall P,Q,x,y: m(P,x,y) \leftarrow
m(Q,x,y)$. Encapsulation further maps each predicate symbol $p \in \mathcal{P}$
to a new constant $p \in \mathcal{C}$ \footnote{Such sly cheating of FOL
semantics is enabled by Prolog where $\mathcal{P,C}$ need not be disjoint.}.

\section{Framework}
\label{sec:Framework}

\subsection{Metarule languages}
\label{subsec:Metarule languages}

\begin{table}[t]
\centering
	\begin{tabular}{lll}
		\textbf{Name}	& \textbf{Quantification}	& \textbf{Metarule} \\
		\toprule
		\emph{Identity} 	& $\exists_{\in \mathcal{P}} P,Q, \forall_{\in \mathcal{C}}x,y$ & $P(x,y) \leftarrow Q(x,y)$\\ % Identity
		\emph{Inverse}  	& $\exists_{\in \mathcal{P}} P,Q, \forall_{\in \mathcal{C}}x,y$ & $P(x,y) \leftarrow Q(y,x)$\\ % Inverse
		\emph{XY-XY-XY}         & $\exists_{\in \mathcal{P}} P,Q,R, \forall_{\in \mathcal{C}}x,y$ & $P(x,y) \leftarrow Q(x,y), R(x,y)$\\
		\emph{XY-XY-YX}         & $\exists_{\in \mathcal{P}} P,Q,R, \forall_{\in \mathcal{C}}x,y$ & $P(x,y) \leftarrow Q(x,y), R(y,x)$\\
		\emph{Chain} 		& $\exists_{\in \mathcal{P}} P,Q,R, \forall_{\in \mathcal{C}}x,y,z$ & $P(x,y) \leftarrow Q(x,z), R(z,y)$\\ % Chain
		\emph{XY-XZ-YZ}         & $\exists_{\in \mathcal{P}} P,Q,R, \forall_{\in \mathcal{C}}x,y,z$ & $P(x,y) \leftarrow Q(x,z), R(y,z)$\\ % Switch
		\emph{XY-YX-XY}         & $\exists_{\in \mathcal{P}} P,Q,R, \forall_{\in \mathcal{C}}x,y$ & $P(x,y) \leftarrow Q(y,x), R(x,y)$\\
		\emph{XY-YX-YX}         & $\exists_{\in \mathcal{P}} P,Q,R, \forall_{\in \mathcal{C}}x,y$ & $P(x,y) \leftarrow Q(y,x), R(y,x)$\\
		\emph{XY-YZ-XZ}         & $\exists_{\in \mathcal{P}} P,Q,R, \forall_{\in \mathcal{C}}x,y,z$ & $P(x,y) \leftarrow Q(y,z), R(x,z)$\\
		\emph{XY-YZ-ZX}         & $\exists_{\in \mathcal{P}} P,Q,R, \forall_{\in \mathcal{C}}x,y,z$ & $P(x,y) \leftarrow Q(y,z), R(z,x)$\\
		\emph{XY-ZX-YZ}         & $\exists_{\in \mathcal{P}} P,Q,R, \forall_{\in \mathcal{C}}x,y,z$ & $P(x,y) \leftarrow Q(z,x), R(y,z)$\\
		\emph{XY-ZX-ZY}         & $\exists_{\in \mathcal{P}} P,Q,R, \forall_{\in \mathcal{C}}x,y,z$ & $P(x,y) \leftarrow Q(z,x), R(z,y)$\\ % Swap
		\emph{XY-ZY-XZ}         & $\exists_{\in \mathcal{P}} P,Q,R, \forall_{\in \mathcal{C}}x,y,z$ & $P(x,y) \leftarrow Q(z,y), R(x,z)$\\
		\emph{XY-ZY-ZX}         & $\exists_{\in \mathcal{P}} P,Q,R, \forall_{\in \mathcal{C}}x,y,z$ & $P(x,y) \leftarrow Q(z,y), R(z,x)$\\
		\bottomrule
	\end{tabular}
\caption{``Canonical" set of fully-connected $H^2_2$ metarules.}
\label{tab:canonical_h22_metarules}
\end{table}

In this section we introduce a formal notation for sets of metarules.

\begin{definition} A metarule language $\mathscr{M}^{l}_{a}$ is a set of
	metarules and their instances where $l$ is a natural number or an
	interval over the natural numbers, denoting the number of literals in
	clauses in $\mathscr{M}^{l}_{a}$, and $a$ is a natural number, or an
	interval over the natural numbers, or a sequence of natural numbers,
	denoting the arities of literals in clauses in $\mathscr{M}^{l}_{a}$.
\label{def:languages}
\end{definition}

When the arity term, $a$, in $\mathscr{M}^{l}_{a}$ is a sequence, a total
ordering is assumed over the literals in each clause in $\mathscr{M}^{l}_{a}$
such that a) the positive literal is ordered before any negative literals, b)
negative literals are ordered by lexicographic order of the names of their
symbols and variables and c) literals with the same symbol and variable names
are ordered by ascending arity. 

\begin{example} $\mathscr{M}^{3}_{2}$ is the language of metarules and
	first-order definite clauses with exactly three literals each of arity
	exactly 2; $\mathscr{M}^{[2,3]}_{\,[1,2]}$ is the language of metarules
	and first-order definite clauses with 2 or 3 literals of arities between
	1 and 2; and $\mathscr{M}^{3}_{\langle 1,2,3 \rangle}$ is the language
	of metarules and first-order definite clauses having exactly three
	literals, a positive literal of arity 1 and two negative literals of
	arity 2 and arity 3, in that order.
\end{example}

The arities of literals in third-order metarules may not be known or relevant,
in which case the arity term may be omitted from the formal notation of a
language.

\begin{example} $\mathscr{M}^3$ is the language of third-order metarules,
	second-order metarules and first-order definite clauses with exactly 3
	literals. $\mathscr{M}^{[1,5]}$ is the language of third- and
	second-order metarules and first-order definite clauses with 1 to 5
	literals.
\end{example}

Of special interest to MIL is the $\mathscr{M}^{[2,3]}_{2}$ language of
fully-connected (see Definition \ref{def:fully_connected}) second-order
metarules with one function symbol which is decidable when $\mathcal{P}$ and
$\mathcal{C}$ are finite \citep{Muggleton2015} and second-order variables are
\emph{universally} quantified by first-order encapsulation. We denote this
language exceptionally as $H^2_2$ in keeping with the MIL literature. The set of
14 second-order $H^2_2$ metarules defined in \cite{Cropper2015} are given in
Table \ref{tab:canonical_h22_metarules}.

\subsection{Metarule taxonomy}
\label{subsec:Metarule taxonomy}

\begin{table}[t]
\centering
	\begin{tabular}{lll}
		\textbf{Name}		& \textbf{Quantification}							& \textbf{Metarule} \\
		\toprule
		\emph{Meta-Monadic}	& $\exists_{\in \mathcal{P}} P,Q, \forall_{\in \mathcal{C}}x,y,z,u$  		& $P(x,y) \leftarrow Q(z,u)$ \\
		\emph{Meta-Dyadic}	& $\exists_{\in \mathcal{P}} P,Q,R, \forall_{\in \mathcal{C}}x,y,z,u,v,w$  	& $P(x,y) \leftarrow Q(z,u), R(v,w)$ \\
		\bottomrule
	\end{tabular}
\caption{Second-order matrix metarules in the $H^2_2$ language.}
\label{tab:most_general_H22_metarules}
\end{table}

\begin{table}[t]
\centering
	\begin{tabular}{lll}
		\textbf{Name}	& \textbf{Quantification}		& \textbf{Metarule} \\
		\toprule
		\emph{TOM-2}	& $\exists_{\in \mathcal{A}} P,Q$	& ${P} \leftarrow {Q}$ \\
		\emph{TOM-3}	& $\exists_{\in \mathcal{A}} P,Q,R$	& ${P} \leftarrow {Q, R}$ \\
		\bottomrule
	\end{tabular}
\caption{Third-order punch metarules in the $H^2_2$ language.}
\label{tab:third_order_metarules}
\end{table}

Metarules found in the MIL literature, such as the ones listed in Table
\ref{tab:canonical_h22_metarules} are typically \emph{fully-connected}.
Definition \ref{def:fully_connected} extends the definition of fully-connected
metarules in \cite{Cropper2015} to encompass first-order clauses.

\begin{definition} [Fully-connected datalog] \label{def:fully_connected} Let $M$
	be a second-order metarule or a first-order definite clause. Two
	literals $l_i, l_j \in M$ are connected if they share a first-order
	term, or if there exists a literal $l_k \in M$ such that $l_i, l_k$ are
	connected and $l_j, l_k$ are connected.  $M$ is fully connected iff each
	literal $l$ in $M$ appears exactly once in $M$ and $l$ is connected to
	every other literal in $M$.
\end{definition}

%Fully-connected metarules avoid \emph{irrelevancies} in the sense that the truth
%value of a metarule instance depends on the relative values of all its
%universally quantified variables. Conversely, the truth value of instances of a
%metarule such as $P(x) \leftarrow Q(x,z)$ is independent of the relation between
%the values of $x,z$. Additionally fully-connected metarules are free of
%repeating literals, thereby avoiding \emph{redundancies} in the form of
%tautologies. %We leave the proof of irredundancy and relevance of
%%fully-connected metarules for future work.

Fully-connected metarules are specialised, in the sense that all their
universally quantified variables are shared between their literals;
existentially quantified variables may also be shared. Accordingly,
fully-connected metarules can be generalised by replacing each instance of each
of their variables with a new, unique variable of the same order and
quantification as the replaced variable. Applying this generalisation procedure
to the metarules in Table \ref{tab:canonical_h22_metarules} we obtain the
metarules in Table \ref{tab:most_general_H22_metarules}.

We observe that each metarule in Table \ref{tab:most_general_H22_metarules} is a
generalisation of a metarule in the $H^2_2$ language listed in Table
\ref{tab:canonical_h22_metarules} and so the metarules in Table
\ref{tab:most_general_H22_metarules} are the most-general metarules in $H^2_2$.
Further, those most-general $H^2_2$ metarules can themselves be generalised to
the third-order metarules in Table \ref{tab:third_order_metarules} by replacing
each of their literals with a third-order variable. This observation informs our
definition of three taxa of metarules and a total ordering by
$\theta$-subsumption of third- and second-order metarules and their first-order
instances.

\subsection{Punch, sort and matrix metarules}
\label{subsec:Punch, sort and matrix metarules}

In the following sections we employ a moveable type metaphor for the elements of
our metarule taxonomy. In typeset printing, first a \emph{punch} of a glyph is
sculpted in relief in steel. The punch is used to emboss the shape of the glyph
in copper, creating a \emph{matrix}. The copper matrix is filled with molten
soft metal to form a cast of the glyph called a \emph{sort} and used to finally
imprint the glyph onto paper. Thus each ``level" of type elements ``stamps" its
shape onto the next.

Accordingly, in our taxonomy of metarules, a third-order metarule is a
\emph{punch metarule} denoted by $\breve M$, a most-general metarule in a
second-order language is a \emph{matrix metarule} denoted by $\mathring M$, and
a fully-connected second-order metarule is a \emph{sort metarule}, denoted by
$\dot M$. As a mnemonic device the reader may remember that a ``wider" accent
denotes higher generality.

\begin{definition} [Punch metarules] \label{def:punch} A punch metarule $\breve
	M$ is a third-order definite clause of the form: $\exists_{\in
	\mathcal{A}} A_1, A_2, ..., A_n: \{A_1 \vee \neg A_2 \vee ... \vee \neg
	A_n\}$.
\end{definition}

\begin{definition} [Matrix metarules] \label{def:matrix} A matrix metarule
	$\mathring M$ (not to be confused with the linear algebra, or
	first-order logic concepts of a matrix) is a second-order definite
	clause of the form: $\exists_{\in \mathcal{P}} \tau, \exists_{\in
	\mathcal{C}} \sigma, \forall_{\in \mathcal{C}} \rho: \{L_1 \vee \neg L_2
	\vee ... \vee \neg L_n\}$ where $\tau, \sigma, \rho$ are disjoint sets
	of variables, each $L_i \in \mathring M$ is a second-order atom
	$P_i(v_{i1}, ..., v_{im})$, $P_i \in \tau, v_{i1}, ..., v_{im} \in
	\sigma \cup \rho$ and none of $P_i, v_{i1}, ..., v_{im}$ is shared with
	any other literal $L_k \in \mathring M$.
\end{definition}

\begin{definition} [Sort metarules] \label{def:sort} A sort metarule $\dot M$
	(not to be confused with the logic programming concept of a sort) is a
	second-order definite clause of the form: $\exists_{\in \mathcal{P}}
	\tau, \exists_{\in \mathcal{C}} \sigma, \forall_{\in \mathcal{C}} \rho:
	\{L_1 \vee \neg L_2 \vee ... \vee \neg L_n\}$ where $\tau, \sigma, \rho$
	are disjoint sets of variables, each $L_i \in \dot M$ is a second-order
	atom $P_i(v_{i1}, ..., v_{im})$, $P_i \in \tau, v_{i1}, ..., v_{im} \in
	\sigma \cup \rho$ and at least one $L_i \in \dot M$ shares a variable in
	$\tau \cup \sigma \cup \rho$ with at least one other literal $L_k \in
	\dot M$.
\end{definition}

\begin{note} Sort metarules are not necessarily fully-connected, rather
	fully-connected metarules are a sub-set of the sort metarules. The
	metarules typically used in the MIL literature, such as the 14
	Canonical $H^2_2$ metarules in Table \ref{tab:canonical_h22_metarules}
	are \emph{fully connected sort metarules}.
\end{note}

We define metarules as sets of literals according to Section \ref{subsec:Logical
notation}. As discussed at the end of that section we will write clauses in the
logic programming convention, as implications, and this also applies to
metarules. Additionally, in keeping with MIL convention we will write metarules
concisely without quantifiers instead denoting quantification by means of
capitalisation: uppercase letters for existentially quantified variables,
lower-case letters for universally quantified variables.

Thus, we will write a punch metarule $\exists_{\in \mathcal{A}}P,Q,R: \{P \vee
\neg Q \vee \neg R\}$ as an implication $P \leftarrow Q, R$, a matrix metarule
$\exists_{\in \mathcal{P}} P,Q,R, \forall_{\in \mathcal{C}}x,y,z,u,v,w: \{P(x,y)
\vee \neg Q(z,u) \vee \neg R(v,w)\}$ as an implication $P(x,y) \leftarrow
Q(z,u), R(v,w)$ and a sort metarule $\exists_{\in\mathcal{P}}P, Q, R$,
$\exists_{\in \mathcal{C}}X, \forall_{\in \mathcal{C}}x,y,z: \{P(x,y) \vee \neg
Q(x,z) \vee \neg R(X)\}$ as an implication $P(x,y) \leftarrow Q(x,z), R(X)$.

Defining metarules as sets of clauses facilitates their comparison in terms of
generality, while denoting them as implications, without quantifiers, makes them
easier to read and closely follows their implementation in MIL systems as Prolog
clauses (with encapsulation).

\subsection{Metarule generality order}
\label{subsec:Metarule generality order}

We extend $\theta$-subsumption between clauses, as defined by
\cite{Plotkin1972}, to encompass metarules with existentially quantified
variables:

\begin{definition} [Meta-subsumption] Let $C$ be a metarule or a first-order
	definite clause and $D$ be a metarule or a first-order definite clause.
	$C \preceq D$ (read $C$ subsumes $D$) iff $\exists \vartheta, \Theta: C
	\vartheta \Theta \subseteq D$ where $\vartheta$ is a substitution of the
	universally quantified variables in $C$ and $\Theta$ is a
	metasubstitution of the existentially quantified variables in $C$.
\label{def:subsumption}
\end{definition}

\begin{lemma} [3rd-order subsumption] Let $\breve M$ be a punch metarule in the
	language $\mathscr{M}^{l}$ and $\mathring M$ be a matrix metarule in the
	language $\mathscr{M}^{k}_{a}$. Then, $\forall a: l \leq k \rightarrow
	\breve M \preceq \mathring M$.%
\label{lem:punch_matrix_subsumption}
\end{lemma}

\begin{proof} Let $\breve M = \{A_1 \vee \neg A_2 ... \vee \neg A_l\}$ and
	$\mathring M = \{L_1 \vee \neg L_2 ... \vee \neg L_k\}$. Assume a total
	ordering over the literals in $\breve M$ and $\mathring M$ as described
	in Section \ref{subsec:Metarule taxonomy}. Let $\vartheta = \emptyset$
	and let $\Theta$ be the metasubstitution that maps each $A_i \in \breve
	M$ to each $L_i \in \mathring M$. While $l \leq k$, $\exists \vartheta,
	\Theta$ and $\breve M \vartheta \Theta \subseteq \mathring M$.
\end{proof}

\begin{lemma} [2nd-order subsumption] Let $\mathring M$ be a matrix metarule in
	the language $\mathscr{M}^{l}_{a}$ and $\dot M$ be a sort metarule in
	the language $\mathscr{M}^{k}_{b}$, where $a,b$ are two integers or two
	sequences of integers having the same first element. Then $l \leq k
	\rightarrow \mathring M \preceq \dot M$ iff $a = b$ or $a$ is a
	subsequence of $b$.
\label{lem:matrix_sort_subsumption}
\end{lemma}

\begin{proof} Let $P_1, ..., P_l$ be the existentially quantified variables and
	$v_1,...v_n$ the universally quantified variables in $\mathring M \in
	\mathscr{M}^{l}_{a}$. Let $Q_1, ..., Q_k$ be the existentially
	quantified variables and $u_1, ..., u_m$ the universally quantified
	variables in $\dot M \in \mathscr{M}^{k}_{b}$. Assume a total ordering
	over the literals in $\mathring M$ and $\dot M$ as described in Section
	\ref{subsec:Metarule taxonomy}. Let $\vartheta$ be the substitution that
	maps each $v_i$ to $u_i$ and $\Theta$ the metasubstitution that maps
	each $P_j$ to $Q_j$. While $l \leq k$, and either $a \leq b$ or $a$ is a
	subsequence of $b$, $\exists\vartheta, \Theta$ and $\mathring M
	\vartheta \Theta \subseteq \dot M$.
\end{proof}

%\begin{proof} Let $P_1, ..., P_l$ be the existentially quantified variables and
%	$v_1,...v_n$ be the universally quantified variables in $\mathring M \in
%	\mathscr{M}^{l}_{a}$. Let $Q_1, ..., Q_k$ be the existentially
%	quantified variables and $u_1, ..., u_m$ be the universally quantified
%	variables in $\dot M \in \mathscr{M}^{k}_{b}$. Let $\vartheta$ be the
%	substitution that maps each $v_i$ to $u_i$ and $\Theta$ be the
%	metasubstitution that maps each $P_j$ to $Q_j$. Suppose $l = k$ and $a =
%	b$. Then $\exists \vartheta \Theta$ and $\mathring M \vartheta \Theta =
%	\dot M$. Suppose $l < k$ and $a < b$ or $a \subseteq b$. Then $\exists
%	\vartheta \Theta$ and $\mathring M \vartheta \Theta \subset \dot M$.
%	Therefore, while $l \leq k$ and $a = b$ or $a \subseteq b$ $\exists
%	\vartheta \Theta: \mathring M \vartheta \Theta \subseteq \dot M$ and
%	$\mathring M \preceq \dot M$.
%\end{proof}

\begin{lemma} [1st-order subsumption] Let $\dot M$ be a sort metarule in the
	language $\mathscr{M}^{l}_{a}$ and $C$ be a first-order clause in the
	language $\mathscr{M}^{k}_{b}$, where $a,b$ are two integers or two
	sequences of integers having the same first element. Then $l \leq k
	\rightarrow \dot M \preceq C$ iff $a = b$ or $a$ is a subsequence of
	$b$.
\label{lem:sort_clause_subsumption}
\end{lemma}

\begin{proof} Let $P_1, ..., P_l$ be the existentially quantified variables and
	$v_1,...v_n$ be the universally quantified variables in $\dot M \in
	\mathscr{M}^{l}_{a}$. Let $Q_1, ..., Q_k$ be the predicate symbols and
	constants in $C \in \mathscr{M}^{k}_{b}$ and $t_1, ..., t_m$ be the
	first-order terms in $C$. Assume a total ordering over the literals in
	$\dot M$ and $C$ as described in Section \ref{subsec:Metarule
	taxonomy}.. Let $\vartheta$ be the substitution that maps each $v_i$ to
	$t_i$ and $\Theta$ be the metasubstitution that maps each $P_j$ to
	$Q_j$. While $l \leq k$, and either $a \leq b$ or $a$ is a subsequence
	of $b$, $\exists \vartheta, \Theta$ and $\dot M \vartheta \Theta
	\subseteq C$.
\end{proof}

%\begin{proof} Let $P_1, ..., P_l$ be the existentially quantified variables and
%	$v_1,...v_n$ be the universally quantified variables in $\dot M \in
%	\mathscr{M}^{l}_{a}$. Let $Q_1, ..., Q_k$ be the predicate symbols in $C
%	\in \mathscr{M}^{k}_{b}$ and $t_1, ..., t_m$ be the first-order terms in
%	$C$. Let $\vartheta$ be the substitution that maps each $v_i$ to $t_i$
%	and $\mu$ be the metasubstitution that maps each $P_j$ to $Q_j$. Suppose
%	$l = k$ and $a = b$. Then $\exists \vartheta \mu$ and $\dot M \vartheta
%	\mu = C$.  Suppose $l < k$ and $a < b$ or $a \subseteq b$. Then $\exists
%	\vartheta \mu$ and $\dot M \vartheta \mu \subset C$. Therefore, while $l
%	\leq k$ and $a = b$ or $a \subseteq b$, $\exists \vartheta \mu: \dot M
%	\vartheta \mu \subseteq C$ and $\dot M \preceq C$.
%\end{proof}

\begin{corollary} Let $\mathscr{M}^{l}_{a}$ be a metarule language. There exists
	a unique, minimal set of matrix metarules $\mathcal{M}^* = \{\mathring
	M_1, ..., \mathring M_n\} \subseteq \mathscr{M}^{l}_{a}$ such that for
	each sort metarule $\dot M \in \mathscr{M}^{l}_{a}$ $\exists \mathring
	M_i \in \mathcal{M}^*: \mathring M_i \preceq \dot M$. Each $\mathring
	M_i \in \mathcal{M}^*$ can be derived by replacing each variable in any
	single $\dot M \in \mathscr{M}^{l}_{a}$ subsumed by $\mathring M_i$ with
	a new, unique variable.
	\label{cor:minimal_sets}
\end{corollary}

\subsection{Metarule specialisation}
\label{subsec:Metarule specialisation}

We now show how first-order clauses and second-order metarules can be derived by
specialisation of more general metarules. We define two ways to specialise a
metarule or a clause: by variable substitution or introduction of new literals.

\begin{note} In the following definitions and theorem, let $M_1, M_2$ be two
	metarules, or a metarule and a definite clause.
\end{note}

% TODO: "$M_1 \vartheta \Theta$ is a v-specialisation of $M_1$" - that's a bit
% confusing. What we want to say is that $M_2$ is a v-specialisation. 
% TODO: see similar constructs in the following two defs.
\begin{definition} [V-specialisation] Let $\vartheta, \Theta$ be substitutions
	of the universally and existentially quantified variables, respectively,
	in $M_1$ such that $M_1 \vartheta \Theta = M_2$. Then $M_1 \vartheta
	\Theta$ is a variable specialisation, or v-specialisation, of $M_1$, and
	$M_2$ is derivable from $M_1$ by v-specialisation, or $M_1 \vdash_v
	M_2$.
\label{def:v_specialisation}
\end{definition}

\begin{definition} [L-specialisation] Let $L$ be a set of literals such that
	$M_1 \cup L = M_2$.  Then $M_1 \cup L$ is a literal specialisation, or
	l-specialisation, of $M_1$, and $M_2$ is derivable from $M_1$ by
	l-specialisation, or $M_1
	\vdash_l M_2$.
\label{def:l_specialisation}
\end{definition}

\begin{definition} [VL- Specialisation] Let $M_1 \vartheta \Theta$ be a
	v-specialisation of $M_1$, $M_1 \cup L$ be an l-specialisation of $M_1$,
	and $M_1 \vartheta \Theta \cup L = M_2$. Then $M_1 \vartheta \Theta \cup
	L$ is a variable and literal specialisation, or vl-specialisation, of
	$M_1$, and $M_2$ is derivable from $M_1$ by vl-specialisation, or $M_1
	\vdash_{vl} M_2$.
\label{def:v_and_l_specialisation}
\end{definition}

\begin{theorem} [Metarule specialisation] $M_1 \preceq M_2 \rightarrow M_1
	\vdash_{vl} M_2$.
\label{thm:specialisations}
\end{theorem}

\begin{proof} If $M_1 \preceq M_2$ then: a) $\exists \vartheta, \Theta: M_1
	\vartheta \Theta \subseteq M_2$ and b) $\exists L: M_1 \vartheta \Theta
	\cup L = M_2$. (a) follows directly from Definition
	\ref{def:subsumption}. (b) follows from (a) and the subset relation: if
	$M_1 \vartheta \Theta \subseteq M_2$ then $\exists L \in M_2: M_2
	\setminus L = M_1 \vartheta \Theta$ and $M_1 \vartheta \Theta \cup L =
	M_2$. By definitions \ref{def:v_specialisation} and
	\ref{def:l_specialisation}, $M_1 \vartheta \Theta$ is a v-specialisation
	of $M_1$ and $M_1 \vartheta \Theta \cup L$ is an l-specialisation of
	$M_1 \vartheta \Theta$. Therefore, $M_1 \vdash_v M_1 \vartheta \Theta
	\vdash_l M_1 \vartheta \Theta \cup L = M_2$ and so $M_1 \vdash_{vl}
	M_2$.
\end{proof}

\begin{observation} There are two special cases of (a) in the proof of Theorem
	\ref{thm:specialisations} with respect to the set of literals $L$:
	either $M_1 \vartheta \Theta = M_2$ or $M_1 \vartheta \Theta \subset
	M_2$. In the case where $M_1 \vartheta \Theta = M_2$, $L = \emptyset$.
	Otherwise, $L \neq \emptyset$.
\end{observation}

\subsection{MIL as metarule specialisation}
\label{subsec:MIL as metarule specialisation}

\begin{algorithm} [t]
	\caption{Resolution-based MIL clause construction}
	\label{alg:mil}
	\textbf{Input}: 1st- or 2nd- order literal $e$; $B^*$, $\mathcal{M}$, elements of a MIL problem.\\
	\textbf{Output}: $M\Theta$, a first-order instance of metarule $M \in \mathcal{M}$.
	\begin{algorithmic}[1]
		\Procedure{Construct}{$\neg e,B^*,\mathcal{M}$}
			\State Select $M \in \mathcal{M}$
			\If{$\exists \sigma, \Sigma: head(M \sigma \Sigma) = e$}
				\If{$\exists  \vartheta, \Theta \supseteq \sigma \Sigma: \neg body(M \vartheta \Theta) \cup B^* \cup \mathcal{M} \vdash_{SLD} \square$} 
					\State \textbf{Return} $M\Theta$
				\EndIf
			\EndIf
		\State \textbf{Return} $\emptyset$
		\EndProcedure
	\end{algorithmic}
\end{algorithm}

In this section we explain MIL as vl-specialisation of metarules.

Algorithm \ref{alg:mil} lists the MIL specialisation operator used in a MIL meta
- interpreter to construct new clauses by refutation of a literal, $\neg e$.
MIL systems implement Algorithm \ref{alg:mil} idiosyncratically: In Metagol $B^*
= B \cup H$ and $\neg e$ is refuted with $B$, $H$ or $\mathcal{M}$ successively
\citep{Cropper2016}; in Louise $B^* = B \cup E^+$ and $\neg e$ is refuted with
$B^* = B \cup E^+$ and $\mathcal{M}$ simultaneously \citep{Patsantzis2021a}.

Initially, $e$ is a positive example in $E^+$ and if refutation of $\neg e$
succeeds the returned clause $M\Theta$ is a clause in the definition of a target
predicate.

If refutation fails, each atom in $body(M\sigma\Sigma)$ in line 4 in Procedure
\textsc{Construct} becomes the input literal $\neg e$ and is refuted recursively
by resolution with $B^* \cup \mathcal{M}$, until $\square$ is derived. In that
case, $M\Theta$ is a clause in the definition of an invented predicate, thus
predicate invention in MIL is achieved by resolution between metarules. Given
that metarules do not have predicate symbols, when $\neg body(M\sigma \Sigma)$
is successfully refuted, the existentially quantified second-order variable $P$
in $head(M\Theta)$ remains free. Hence, a new predicate symbol in $I$ is
substituted for $P$\footnote{To simplify notation, we omit recursion and
predicate invention in Algorithm \ref{alg:mil} and also Algorithm
\ref{alg:vl_specialisation}, below. See Appendix A for a complete description.}.

If $e$ is in $E^-$ and refutation succeeds, $M\Theta$ is inconsistent and must
be replaced in, or discarded from $H$.

We observe that Algorithm \ref{alg:mil} returns vl-specialisations of metarules
in $\mathcal{M}$.

\begin{theorem} [MIL as metarule specialisation] Let $e, B^*, \mathcal{M}$ be as
	in Algorithm \ref{alg:mil}, $M$ be a fully-connected sort metarule
	selected in line 2 of Procedure \textsc{Construct} and $M\Theta =
	$\textsc{Construct}$(\neg e, B^*, \mathcal{M})$. $M\Theta$ is a
	vl-specialisation of $M$.
	\label{thm:mil_as_specialisation}
\end{theorem}

\begin{proof} Let $\vartheta = L = \emptyset$. By definition
	\ref{def:v_and_l_specialisation} $M\vartheta \Theta \cup L = M \Theta$
	is a vl-specialisation of $M$.
\end{proof}

\subsection{Implicit l-specialisation}
\label{subsec:Implicit l-specialisation}

\begin{table}[t]
	\centering
	\begin{tabular}{l}
		\textbf{Occult specialisations} \\
		\toprule
		$(M_1) P(x,y) \leftarrow Q(x,z), R(z,y)$ \\
		$(M_3) P(x,y) \leftarrow Q(x,z), R(z,u), S(u,y)$ \\
		\midrule
		$C_1 = p(x_1,y_1) \leftarrow q(x_1,z_1), \$1(z_1,y_1)$ \\
		$C_2 = \$1(x_2,y_2) \leftarrow p(x_2,z_2), r(z_2,y_2)$ \\
		$C_3 = p(x_1,y_2) \leftarrow q(x_1,x_2), p(x_2,z_2), r(z_2,y_2)$ \\
		\bottomrule
	\end{tabular}
\caption{
$M_3$ is an l-specialisation and self-resolvent of $M_1$. $C_1,C_2$ are
instances of $M_1$. $C_3$ is an instance of $M_3$ and resolvent of $C_1, C_2$.
If $\{C_1, C_2\} \cup B^* \models e$, then $\{M_3\} \cup B^* \models e$. $\$1$
is an invented predicate symbol in Louise's notation.}
\label{tab:occult_specialisations}
\end{table}

Theorem \ref{thm:mil_as_specialisation} states that Procedure \textsc{Construct}
returns vl-specialisations of metarules when the set of introduced literals,
$L$, is empty. This is a special case of vl-specialisation. What about the
general case, when $L \neq \emptyset$? We conjecture that it is not necessary to
explicitly construct such \emph{non-empty} l-specialisations, because the
v\-/specialisations returned by Procedure \textsc{Construct} suffice to
reconstruct non-empty l\-/specialisations by resolution. Suppose $M_1, M_2 \in
\mathcal{M}$ and $C_1, C_2$ are non-empty v\-/specialisations of $M_1, M_2$,
respectively, such that $\exists e \in E^+, \nexists e \in E^-: \{\neg e\} \cup
\{C_1, C_2\} \cup B^* \vdash \square$. We assume that $C_1, C_2$ can resolve
with each other and that $\{\neg e\} \cup B^* \setminus \{e\} \cup \{C_{i \in
\{1,2\}}\} \nvdash \square$. Then, there exists a resolvent $C_3$ of $C_1, C_2$
such that $\exists e \in E^+, \nexists e \in E^-: \{\neg e\} \cup \{C_3\} \cup
B^* \vdash \square$. This follows from the Resolution Theorem
\citep{Robinson1965}. Moreover, there exists a metarule $M_3$ that is a
resolvent of $M_1, M_2$ and such that $C_3$ is a vl-specialisation of $M_3$
where the set of introduced literals, $L$, is not empty. If so, it should not be
necessary to explicitly derive $M_3$ and $C_3$, given $M_1, M_2$ and Procedure
\textsc{Construct}. \cite{Cropper2015} prove our conjecture for the $H^2_2$
language. We leave a more general proof for future work. Table
\ref{tab:occult_specialisations} illustrates the concept of such \emph{occult
specialisations}.

%The preceding is true for any pair of M₁, M₂ having instances in correct
%hypotheses.

%We formalise the concept of ``hidden" non-empty l-specialisations in Definition
%\ref{def:occult_specialisations}, borrowing notation from \citep{Robinson1965}.
%
%\begin{definition} [Occult specialisations] Let $S = B^* \cup \mathcal{M}$,
%	$R^0(S)$ be the set of clauses in $S$ and their resolvents and $R^n(S) =
%	R(R^{n-1}S)$ for $n \leq 0$. $M$ is an occult specialisation of $M_1,
%	M_2 \in \mathcal{M}$ iff $M \in R^n(S)$ and $M$ is a vl-specialisation
%	of $M_1, M_2$ that is not returned by procedure \textsc{Construct} when
%	$\square \in R^n(S)$.
%\label{def:occult_specialisations}
%\end{definition}

\subsection{Metarule specialisation by MIL}
%\subsection{Replacing user-defined metarules with automatically derived metarules.}
\label{sec:Metarule specialisation by MIL}

\begin{algorithm} [t]
	\caption{Resolution-based MIL vl-specialisation}
	\label{alg:vl_specialisation}
	\textbf{Input}: 1st- or higher-order literal $e$; $B^*$ as in Algorithm \ref{alg:mil}; punch and matrix metarules $\mathcal{M}$.\\
	\textbf{Output}: $\dot M$, a fully connected sort metarule.
	\begin{algorithmic}[1]
		\Procedure{VL-Specialise}{$\neg e,B^*,\mathcal{M}$}
			\State Select $M \in \mathcal{M}$
			\If{$\exists \sigma \Sigma: head(M \sigma \Sigma) = e$}
				\If{$\exists  \vartheta, \Theta \supseteq \sigma \Sigma: \neg body(M \vartheta \Theta) \cup B^* \cup \mathcal{M} \vdash_{SLD} \square$} 
					\If{$M\vartheta \Theta$ is fully-connected}
						\State \textbf{Return} $M$.\textsc{Lift}$(\vartheta \Theta)$
					\EndIf
				\EndIf
			\EndIf
		\State \textbf{Return} $\emptyset$
		\EndProcedure
	\end{algorithmic}
\end{algorithm}

\begin{algorithm}[t]
	\caption{Generalisation of ground substitutions}
	\label{alg:lift}
	\textbf{Input}: $\vartheta \Theta$, ground substitution of universally and existentially quantified variables.\\
	\textbf{Output}: $\vartheta \Theta$, generalised by replacing ground terms with variables.
	\begin{algorithmic}[1]
		\Procedure{Lift}{$\vartheta \Theta$}
			\For{$v_i/t_i \in \vartheta \Theta$}
				\If{$t_i \in \mathcal{C} \wedge \forall v_i$}
					\State Replace each instance of $t_i \in \vartheta \Theta$ with a variable $\forall_{\in \mathcal{C}} \; u_i$.
				\ElsIf{$t_i \in \mathcal{C} \wedge \exists v_i$}
					\State Replace each instance of $t_i \in \vartheta \Theta$ with a variable $\exists_{\in \mathcal{C}} \; w_i$.
				\ElsIf{$t_i \in \mathcal{P}$}
					\State Replace each instance of $t_i \in \vartheta \Theta$ with a variable $\exists_{\in \mathcal{P}} \; P_i$.
				\EndIf
			\EndFor
		\EndProcedure
	\end{algorithmic}
\end{algorithm}

Algorithm \ref{alg:mil} learns first-order definite clauses. Our motivation for
this work is to learn metarules that can replace user-defined metarules.
User-defined metarules are fully-connected sort metarules and chosen so that if
$M$ is a user-defined metarule and $M\Theta$ is a vl-specialisation of $M$
returned by Procedure \textsc{Construct} in Algorithm \ref{alg:mil}, then
$\exists e^+ \in E^+: M\Theta \wedge B^* \models e^+$. Therefore, to replace
user-defined metarules with automatically derived metarules, we must
automatically derive fully-connected sort metarules having vl-specialisations
that entail one or more positive examples in $E^+$ with respect to $B^*$.

We achieve this goal by modifying Algorithm \ref{alg:mil}, as Algorithm
\ref{alg:vl_specialisation}, to generalise the substitutions of both universally
and existentially quantified variables in metarules. Such substitutions are
fully-ground by successful resolution with $B^*$ therefore, in order to produce
metarules rather than first-order clauses, we must replace the ground terms in
those substitutions with new variables. We propose Procedure \textsc{Lift} in
Algorithm \ref{alg:lift} to perform this ``variabilisation" operation. 
% TODO: or just call it a generalisation operation?

\begin{lemma} [Fully-connected lifting] Let $\vartheta \Theta/M$ be a
	meta/substitution of a punch or matrix metarule $M$ and $M\vartheta
	\Theta$ be a fully-connected definite clause. The application of
	\textsc{Lift}$(\vartheta \Theta)$ to $M$, $M.$\textsc{Lift}$(\vartheta
	\Theta)$, is a fully-connected sort metarule.
\label{lem:fully_connected_lifting}
\end{lemma}

\begin{proof} Procedure \textsc{Lift} replaces each occurrence of a ground term
	with the same variable throughout $\vartheta \Theta$ so that if two
	literals $l_i,l_k \in M\vartheta \Theta$ share a ground term,
	$\{l_i,l_k\}.$\textsc{Lift}$(\vartheta \Theta) \in
	M.$\textsc{Lift}$(\vartheta \Theta)$ share a variable. Therefore
	$M\vartheta \Theta$ is fully-connected iff $M.$\textsc{Lift}$(\vartheta
	\Theta)$ is fully-connected.
\end{proof}

\begin{lemma} [Lifting subsumption] Let $\vartheta \Theta/M$ be a
	meta/substitution of a punch or matrix metarule M. Then $M \preceq
	M.$\textsc{Lift}$(\vartheta \Theta) \preceq M\vartheta \Theta$.
\label{lem:lifting_subsumption}
\end{lemma}

\begin{proof} $M \preceq M.$\textsc{Lift}$(\vartheta \Theta)$ by Definition
	\ref{def:subsumption}. Construct a meta/substitution $\sigma \Sigma$ by
	mapping each $w_i$ in $v_i/w_i \in \, $\textsc{Lift}$(\vartheta \Theta)$
	to $t_i$ in $v_i/t_i \in \vartheta \Theta$. $\exists \sigma \Sigma:
	M.$\textsc{Lift}$(\vartheta \Theta)\sigma \Sigma = M \vartheta \Theta$,
	therefore $M.$\textsc{Lift}$(\vartheta \Theta) \preceq M \vartheta
	\Theta$.
\end{proof}

\begin{example} Let $M = P(x,y) \leftarrow Q(z,u)$, $\vartheta \Theta = \{P/p,
	Q/q, x/a, y/b, z/a, u/b\}$. Then: $M \vartheta \Theta = p(a,b)
	\leftarrow q(a,b)$, \textsc{Lift}$(\vartheta \Theta) = $ $\{P/P_1,
	Q/Q_1, x/x_1, y/y_1, z/x_1$, $u/y_1\}$, $M.$\textsc{Lift}$(\vartheta
	\Theta) = $ $P_1(x_1,y_1) \leftarrow Q_1(x_1,y_1)$, $\sigma \Sigma = 
	\{P_1/p, Q_1/q, x_1/a, y_1/b \}$ and $M.$\textsc{Lift}$(\vartheta
	\Theta) \sigma \Sigma = p(a,b) \leftarrow q(a,b) = M \vartheta \Theta$.
\end{example}

\begin{theorem} [Soundness] Let $e, B^*, \mathcal{M}$ be as in Algorithm
	\ref{alg:vl_specialisation}. If $M' = $ \textsc{VL- Specialise} $(\neg
	e, B^*, \mathcal{M})$ then $M'$ is a fully-connected sort metarule and
	$\exists \Sigma/ M': M'\Sigma \wedge B^* \models e$.
\label{thm:soundness}
\end{theorem}

\begin{proof} Assume Theorem \ref{thm:soundness} is false. Then, $M' = $
	\textsc{VL-Specialise}$(\neg e, B^*, \mathcal{M})$ and a) $M'$ is not a
	fully-connected sort metarule, or b) $\nexists \Sigma/M': M'\Sigma
	\wedge B^* \models e$. In Procedure \textsc{VL-Specialise} $M' =
	M.$\textsc{Lift}$(\vartheta \Theta)$ is returned iff c) $M$ is the
	punch or matrix metarule selected in line 2, d) $\exists \vartheta,
	\Theta/M: M\vartheta \Theta \cup B^* \cup \{\neg e\} \vdash_{SLD}
	\square$ iff $M\vartheta \Theta \wedge B^* \models e$ and e) $M
	\vartheta \Theta$ is a fully-connected definite clause. By Lemma
	\ref{lem:lifting_subsumption}, if (c) and (d) hold then $M' \preceq M
	\vartheta \Theta$ because $\exists \sigma \Sigma: M' \sigma \Sigma = M
	\vartheta \Theta$. Therefore if (c) and (d) hold then $M'\sigma \Sigma
	\wedge B^* \models e$ and $M' \Sigma \wedge B^* \models e$. By Lemma
	\ref{lem:fully_connected_lifting}, if (e) holds then $M'$ is a
	fully-connected sort metarule. Therefore, either $M'
	\neq$\textsc{VL-Specialise}$(\neg e, B^*, \mathcal{M})$ and (a), (b) are
	false, or Theorem \ref{thm:soundness} is true. This refutes the
	assumption and completes the proof.
\end{proof}

\subsection{Cardinality of metarule languages}
\label{subsec:Cardinality of metarule languages}

% TODO: say something about the fact that we restrict attention to languages
% derived from punches of length k, to allow for polynomial-time enumeration.
In this section we turn our attention to the cardinalities of metarule languages
and show that they are polynomial in the number of punch metarule literals.

\begin{table}[t]
\centering
	\begin{tabularx}{\columnwidth}{lX}
		\textbf{Language} & \textbf{Cardinality} \\
		\toprule
		$\breve M \in \mathscr{M}^{k}$ & $k$ \\
		$\mathring M \in \mathscr{M}^{k}$ & $\leq k(a^k/k!)$ \\
		$\dot M \in \mathscr{M}^{k}$  & $< (2n-1)^n/n!$ \\
		$\Theta/M \in \mathscr{M}^{k}$ & $< p^kc^n$ \\
		Ground definite clauses $C \in \mathscr{M}^{k}$ & $< c^n$ \\
		\midrule
		%Cardinality of $\mathscr{M}^{k}$ & $\leq \sum_{i=1}^{k} i(a^i/i!) ((2n-1)^n/n!) p^ic^n c^n$ \\ \\
		Cardinality of $\mathscr{M}^{k}$ & $\leq \sum_{i=1}^{k} \dfrac{ia^i (2n-1)^n p^ic^{2n}}{i!n!}$ \\
		\bottomrule
	\end{tabularx}
\caption{Cardinality of metarule languages with at most $k$ body literals
according to Lemmas \ref{lem:punch_complexity} -
\ref{lem:ground_clause_complexity} and Theorem
\ref{thm:metarule_language_complexity}. $k$: maximum number of body literals in
punch metarules; $a$: cardinality of the set, $\mathcal{A}_{\mathring M}$,
of matrix metarule literals; $n$: (constant) number of existentially and
universally quantified variables in sort metarule literals; $c$: cardinality of
the set, $\mathcal{C}$, of constants; $p$: cardinality of the set,
$\mathcal{P}$, of predicate symbols.}
\label{tab:metarule_language_complexity}
\end{table}

\begin{definition} [Clause length] Let $C$ be a metarule or a first-order
	definite clause. The length of $C$ is the number of literals in $C$. 
\end{definition}

\begin{note} In the following lemmas and proofs, metarules that differ only in
	the names of their variables are considered identical.
	\label{not:renaming}
\end{note}

% Cardinality of punch languages
\begin{lemma} [Number of punch metarules] The number of punch metarules of
	length in $[1,k]$ is $k$.
	\label{lem:punch_complexity}
\end{lemma}

\begin{proof} There exist $k$ n-tuples of third-order variables for $n \in
	[1,k]$. Exactly one definite clause can be formed from each such
	n-tuple (see Note \ref{not:renaming}).
\end{proof}

\begin{example} Suppose $k = 3$. The set of n-tuples of third-order variables in
	punch metarules of length 1 to $k$ is $\{\{P\}, \{P,Q\}, \{P,Q,R\}\}$.
	Exactly one definite clause can be formed from each such n-tuple:
	$\{P\}$, $\{P \vee \neg Q\}$ and $\{P \vee \neg Q \vee \neg R\}$.
\end{example}

% Cardinality of matrix languages
\begin{lemma} [Number of matrix metarules] Let $\mathcal{A}_{\mathring M}
	\subseteq \mathcal{A}$ be the set of matrix metarule atoms and $a =
	|\mathcal{A}_{\mathring M}|$. The number of matrix metarules of length
	$k$ is at most $k(a^k/k!)$.%
\label{lem:matrix_complexity}
\end{lemma}

\begin{proof} Let $\{A_1,...,A_k: A_i \in \mathcal{A}_{\mathring M}\}$ be the
	k-tuple of atoms in a matrix metarule of length $k$. The number of
	such k-tuples is the number of subsets of $\mathcal{A}_{\mathring M}$ of
	length $k$, called the k-combinations of $\mathcal{A}_{\mathring M}$,
	which is equal to the binomial coefficient $\binom{a}{k}$, which is at
	most $a^k/k!$ for $1 \leq k \leq a$ \citep{Cormen2001}. Note that if $a$
	is less than $k$ a matrix metarule of length $k$ cannot be formed
	because matrix metarule literals must be distinct atoms. For each such
	k-tuple, $T$, exactly $k$ definite clauses can be formed by taking one
	atom in $T$ as the single positive literal, in turn.
\end{proof}

\begin{example} Suppose $k = 3$, $\mathcal{A}_{\mathring M} = \{P(x,y),
	Q(z,u,v), R(w)\}$. The set, $T$, of 3-tuples of atoms in
	$\mathcal{A}_{\mathring M}$ is $\{ \{P(x,y), Q(z,u,v)$, $R(w) \} \} =
	\{\mathcal{A}_{\mathring M}\}$. The set of definite clauses formed by
	taking each atom in a 3-tuple in $T$ as a positive literal in turn is:
	$\{\{P(x,y)$, $\neg Q(z,u,v)$, $\neg R(w)\}$, $\{\neg P(x,y)$,
	$Q(z,u,v)$, $\neg R(w)\}$, $\{\neg P(x,y)$,$\neg Q(z,u,v)$, $R(w)\}\}$.
\end{example}

% Cardinality of sort languages
\begin{lemma} [Number of sort metarules] Let $e$ be the number of existentially
	quantified first- and second-order variables, and $u$ be the number of
	universally quantified first-order variables, in all sort metarules of
	length $k$ in the language $\mathscr{M}^{k}_{b}$. Let $n = e + u$. The
	number of sort metarules in $\mathscr{M}^{k}_{b}$ is less than
	$(2n-1)^n/ n!$.%
	\label{lem:sort_complexity}
\end{lemma}

\begin{proof} Let $\{P_1, ..., P_e, v_1, ..., v_u\}$ be the
	multiset\footnote{Informally, a multiset is a collection of a set's
	elements each repeating a number of times equal to its multiplicity.}
	with cardinality $n = e + u$ of existentially quantified first- and
	second-order variables $P_i$ and universally quantified first-order
	variables $v_j$ each with multiplicity 1 or more. Let $S$ be the set of
	all such multisets. Let $S'$ be the set of multisets in $S$ each
	containing existentially quantified second-order variables of total
	multiplicity $k$, existentially quantified first-order variables of
	total multiplicity $e-k$, and universally quantified first-order
	variables with total multiplicity $u$ at least one of which has
	multiplicity between 2 and u. Multiplicities of variables in elements of
	$S'$ are constrained by Definition \ref{def:sort} and the cardinality of
	elements of $S'$ is $n = e + u$ therefore $S'$ is the set of multisets
	of variables in sort metarules of length $k$ in the language
	$\mathscr{M}^{k}_{b}$. The cardinality of $S$ is equal to the multiset
	coefficient $\left(\!\binom{n}{n}\!\right)$ which is equal to the
	binomial coefficient $\binom{2n-1}{n}$ \citep{Stanley1997}. $S$
	necessarily includes elements not in $S'$, for example multisets
	containing no universally quantified variables with multiplicity 2.
	Therefore $|S'| < |S|$ and so $|S'| < \binom{2n-1}{n}$. $(2n-1)^n/n!$ is
	an upper bound for $\binom{2n-1}{n}$ for $1 \leq n \leq 2n-1$ which is
	always the case, therefore $|S'| < (2n-1)^n/n!$.
\end{proof}

\begin{example} Suppose $k = 3$, n = 9. The set, $S$, of n-multisets of
	existentially and universally quantified variables is $\{$ $\{P, P, P,
	x$, $x, x, x$, $x, x\}$,  $\{P, P, P, x, x, x, x, x, y, \}$, $\{P, P, P,
	x, x, x, x, x, z\}$,  $... \}$ etc. The set, $S'$, of n-multisets of
	existentially and universally quantified variables in sort metarules of
	length 3 in the language $\mathscr{M}^{3}_{2}$ is $\{ \{P, P, P, x, x,
	x, x, x, x\},  \{P, P, P$, $y, y, x, x, x, x\}, \{P, P, P, y, x, y, x,
	x, x\},  \{P, P, P$, $y, x, x, y, x, x\},  ... \}$ etc.
\end{example}

% Cardinality of sets of metasubstitutions
\begin{lemma} [Number of metasubstitutions] Let $p = \mathcal{P}$, $c =
	|\mathcal{C}|$ and let $n$ be as in Lemma \ref{lem:sort_complexity}. The
	number of metasubstitutions of sort metarules of length $k$ is less than
	$p^kc^n$.
	\label{lem:metasub_complexity}
\end{lemma}

\begin{proof} Let $h$ be the number of predicate symbols in the heads and $b$
	the number of predicate symbols in literals in the body, of all
	metasubstitutions of a sort metarule of length $k$, and let $e$ be as in
	Lemma \ref{lem:sort_complexity}. Let
	$\{H,B_1,...,B_{k-1},c_1,...,c_{e-k}\}$ be the e-tuple of a predicate
	symbol $H$ substituting the existentially quantified second-order
	variable in the head literal, predicate symbols $B_i$ substituting
	existentially quantified second-order variables in the body literals,
	and constants $c_j$ substituting existentially quantified first-order
	variables in all literals, in a sort metarule of length $k$.  There
	exist $hb^{k-1}c^{e-k}$ such e-tuples. $hb^{k-1}$ is at most $p^k$ as
	when all symbols in $\mathcal{P}$ are of target predicates. $c^{e-k}$ is
	always less than $c^n$ because at least $k$ existentially-quantified
	variables in a sort metarule of length $k$ must be second-order.
	Therefore, $hb^{k-1}c^{e-k}$ is less than $p^kc^n$.
\end{proof}

\begin{example} Suppose $k = 3, e = 4$, $H \in \{p\}$, $B_i \in \{q,r\}$,
	$\mathcal{C}$ $ = \{a,b,c\}$. The set of 4-tuples of predicate symbols
	and constants in metasubstitutions of sort metarules of length 3 with
	one existentially quantified first-order variable is: $\{ \{p,q,q,a\}$,
	$\{p,q,q,b\}$, $\{p,q,q,c\}$, $\{p,q$,$r,a\}$, $...$, $\{p,r,r,a\}$,
	$\{p,r,r,b\}$, $\{p,r,r,c\}\}$.
\end{example}

\begin{observation} Lemma \ref{lem:metasub_complexity} is a refinement of
	earlier results by \cite{Lin2014, Cropper2018} who calculate the
	cardinality of the set of metasubstitutions of a single sort metarule as
	$p^k$ (or $p^3$ for $H^2_2$ metarules). Our result takes into account,
	firstly the restriction that only symbols in $E^+$ and $I$ can be substituted
	for second-order variables in the heads of sort metarules, and secondly
	the possible metasubstitution of existentially quantified first-order
	variables by constants, neither of which is considered in the earlier
	results.
	\label{obs:earlier_results}
\end{observation}

%% Cardinality of sets of fully-ground clauses.
\begin{lemma} [Number of ground clauses] Let $n,c$ be as in Lemmas
	\ref{lem:sort_complexity} and \ref{lem:metasub_complexity}. The number
	of ground substitutions of the universally quantified variables in a
	sort metarule is less than $c^n$.
	\label{lem:ground_clause_complexity}
\end{lemma}

\begin{proof} Let $\{v_1,...,v_c\}$ be the c-tuple of constants substituted for
	$u$ universally quantified variables. There are $c^u$ such u-tuples. In
	a ground substitution of the universally quantified variables in a sort
	metarule, $c^u$ is always less than $c^n$ because $n = e+u$, where $e$
	is as in Lemma \ref{lem:sort_complexity}, and there are exactly $k > 0$
	existentially quantified second-order variables in a sort metarule of
	length $k$, therefore $e > 0$.
\end{proof}

\begin{example} Let $u = 3,\; \mathcal{C} = \{a,b,c\}$. The set of 3-tuples of
	constants substituted for 3 universally quantified variables is
	$\{\{a,a,a\}$, $\{a,a,b\}$, $\{a,a,c\}$, $\{a,b,a\}$, $..., \{c,c,b\},
	\{c,c,c\} \}$.
\end{example}

% Note on simplifying notation
\begin{note} While we have derived exact results in the proofs of Lemmas
	\ref{lem:metasub_complexity} and \ref{lem:ground_clause_complexity} we
	have chosen to state these two Lemmas in terms of upper bounds in the
	interest of simplifying notation, particularly the notation of Theorem
	\ref{thm:metarule_language_complexity}.
\end{note}

% Cardinality of metarule languages
\begin{theorem} [Cardinality of metarule languages]
	\label{thm:metarule_language_complexity} Let $k,a,n,p,c$ be as in
	Lemmas \ref{lem:punch_complexity} - \ref{lem:ground_clause_complexity}.
	The number of vl- specialisations of punch metarules of length in
	$[1,k]$ is at most: 
	\begin{equation*}
		\sum_{i=1}^{k} \dfrac{ia^i (2n-1)^n p^ic^{2n}}{i!n!}
	\end{equation*}
\end{theorem}

\begin{proof} By Lemma \ref{lem:punch_complexity} there are $k$ punch metarules
	in the language $M^{k}_{}$. The cardinality of the set of
	vl\-/specialisations of the $k$ punch metarules in the language
	$M^{k}_{}$ is the sum of the cardinalities of the sets of
	vl\-/specialisations of punch metarules in each language $M^{i}_{}$,
	where $i \in [1,k]$. 

	Let $i \in [1,k]$. By Lemma \ref{lem:matrix_complexity} there
	exist at most $i(a^i/i!)$ matrix metarule specialisations of a punch
	metarule with $i$ body literals. By Lemma \ref{lem:sort_complexity},
	there exist fewer than $(2n-1)^n/n!$ sort metarule specialisations of
	each such matrix metarule. By Lemma \ref{lem:metasub_complexity} there
	exist fewer than $p^kc^n$ metasubstitutions of each such sort metarule.
	By Lemma \ref{lem:ground_clause_complexity} there exist fewer than
	$c^n$ ground first-order clause specialisations of each such
	metasubstitution.

	Thus, the cardinality of the set of vl-specialisations of punch
	metarules in the language $M^{k}_{}$ is at most the sum for all $i \in
	[1,k]$ of the product $i(a^i/i!) ((2n-1)^n/ n!) p^ic^nc^n$. We may
	rewrite this product as the fraction $\frac{ia^i (2n-1)^n
	p^ic^{2n}}{i!n!}$. 
\end{proof}

\begin{corollary} Each metarule language $\mathscr{M}^{l}_{a}$ is enumerable in
	time polynomial to the number of literals in the most general metarule
	in $\mathscr{M}^{l}_{a}$, i.e. $l$.
\end{corollary}

\section{Implementation}
\label{sec:Implementation}

% TODO: you need to say something about using Louise and what it does or why
% you choose it

\begin{algorithm} [t]
	\caption{TOIL-2 look-ahead heuristic}
	\label{alg:look_ahead}
	\textbf{Input}: Non-ground metarule instance $M\vartheta \Theta$; substitution buffer $S = \{c_1 \mapsto k_1, ..., c_n \mapsto k_n\}$. \\
	\textbf{Output}: True if $M \vartheta \Theta$ can be fully-connected given $S$; else false.
	\begin{algorithmic}[1]
		\Procedure{Look-Ahead}{$M \vartheta \Theta = \{P_1(t_{11},...,t_{1k}), ..., P_n(t_{n1},...,t_{nj})\}, S$}
			\State $F \Leftarrow \{t_{mi}: free\_variable(t_{mi}) \wedge t_{mi} \in L \wedge L \in M \vartheta \Theta\}$
			\State $C \Leftarrow \{c \mapsto 1: c \mapsto 1 \in S\}$
			\If{$|C| \leq |F|$}
				\State \Return true
			\EndIf
			\State \Return false
		\EndProcedure
	\end{algorithmic}
\end{algorithm}

We have created a prototype, partial implementation of Algorithms
\ref{alg:vl_specialisation} and \ref{alg:lift} in Prolog, as a new module added
to Louise\footnote{Our new module is available from the Louise repository, at
the following url:
\url{https://github.com/stassa/louise/blob/master/src/toil.pl}}. The
implementation is partial in that it performs only v-specialisation of punch and
matrix metarules, but not l-specialisation. For clarity, we will refer to this
new module as TOIL (an abbreviation of \emph{Third Order Inductive Learner}). We
now briefly discuss TOIL but leave a full description for future work, alongside
a complete implementation\footnote{We reserve the title \emph{TOIL: A full-term
report} for this future work.}.

We distinguish punch metarule specialisation in TOIL as TOIL-3 and matrix
metarule specialisation as TOIL-2. Both sub-systems are implemented as variants
of the Top Program Construction algorithm in Louise. Each subsystem takes as
input a MIL problem with punch or matrix metarules, respectively for TOIL-2 and
TOIL-3, instead of sort metarules, and outputs a set of sort metarules.

According to line 5 of Procedure \textsc{VL-Specialise} in Algorithm
\ref{alg:vl_specialisation}, both sub-systems test that a ground instance $M
\vartheta \Theta$ of an input metarule $M$ is fully\-/connected before passing
it to their implementation of Procedure \textsc{Lift}. To do so, TOIL-2
maintains a ``substitution buffer", $S$, of tuples $c \mapsto k$ where each $c$
is a constant and each $k$ is the number of first-order variables in $M$
substituted by $c$. If, when line 5 is reached, $S$ includes any tuples where
$k = 1$, $M \vartheta \Theta$ is not fully\-/connected. $S$ is first
instantiated to the constants in an input example. When a new literal $L$ of $M$
is specialised, the first-order variables in $L$ are first substituted for
constants in $S$, ensuring that $L$ is connected to literals earlier in $M$. $L$
is then resolved with $B^*$. If resolution succeeds, a ``look-ahead" heuristic,
listed in Algorithm \ref{alg:look_ahead}, attempts to predict whether the now
fully-ground $L$ allows a fully\-/connected instantiation of $M$ to be derived.
If so, $S$ is updated with the new constants derived during resolution and the
new counts of existing constants. If not, the process backtracks to try a new
grounding of $L$.

TOIL-3 restricts instantiation of punch metarule literals to the set
$\mathcal{A}_{\mathring M, B}$, of matrix metarule literals unifiable with the
heads of clauses in $B^*$ ($\mathcal{A}_{\mathring M, B}$ is generated
automatically by TOIL-3). Because atoms in $\mathcal{A}_{\mathring M, B}$ are
non-ground, it is not possible to apply the look-ahead heuristic employed in
TOIL-2; TOIL-3 only uses the substitution buffer to ensure derived metarules are
fully\-/connected.

Theorem \ref{thm:metarule_language_complexity} predicts that metarule languages
are enumerable in polynomial time, but generating an entire metarule language is
still expensive---and unnecessary. To avoid over-generation of metarule
specialisations, TOIL limits the number of attempted metarule specialisations,
in three ways: a) by sub-sampling, i.e. training on a randomly selected sample
of $E^+$; b) by directly limiting the number of metarule specialisation
attempts; and c) by a cover-set procedure that removes from $E^+$ each example
entailed by the last derived specialisation of an input metarule, before
attempting a new one.

TOIL cannot directly derive sort metarules with existentially quantified
first-order variables. These must be simulated by monadic background predicates
representing possible theory constants, e.g., $pi(3.14)$, $e(2.71)$, $g(9.834)$,
$c(300000)$, etc.

We leave a formal treatment of the properties of the connectedness constraints
and specialisation limits described above to the aforementioned future work.

\section{Experiments}
\label{sec:Experiments}

A common criticism of the MIL approach is its dependence on user-defined
metarules. In this Section we show experimentally that automatically derived
fully-connected sort metarules can replace user-defined fully-connected sort
metarules as shown in Section \ref{sec:Metarule specialisation by MIL}, thus
addressing the aforementioned criticism. We formalise our motivation for our
experiments as Experimental Hypotheses \ref{hyp:predictive_accuracy} and
\ref{hyp:running_time}.

\begin{exp-hypothesis} Metarules learned by TOIL can replace user\-/defined
	metarules without decreasing Louise's predictive accuracy.
\label{hyp:predictive_accuracy}
\end{exp-hypothesis}

\begin{exp-hypothesis} Metarules learned by TOIL can replace user\-/defined
	metarules without increasing Louise's training time.%
\label{hyp:running_time}
\end{exp-hypothesis}

\subsection{Experiment setup}
\label{subsec:Experiment setup}

We conduct a set of metarule replacement experiments where an initial set,
$\mathcal{M}$, of user-defined, fully-connected sort metarules are progressively
replaced by metarules learned by TOIL\footnote{Experiment code and datasets are
available from \url{https://github.com/stassa/mlj\_2021}}.

Each metarule replacement experiment proceeds for $k = |\mathcal{M}| + 1$ steps.
Each step is split into three separate legs. We repeat the experiment for $j =
10$ runs at the end of which we aggregate results. Each leg is associated with a
new set of metarules: $\mathcal{M}_1$, $\mathcal{M}_2$ and $\mathcal{M}_3$ for
legs 1 through 3, respectively. At the start of each run we initialise
$\mathcal{M}_1$ to $\mathcal{M}$, and $\mathcal{M}_2$, $\mathcal{M}_3$ to
$\emptyset$. At each step $i$ \emph{after the first}, we select, uniformly at
random and without replacement, a new user-defined metarule $M_i$ and set
$\mathcal{M}_1 = \mathcal{M}_1 \setminus \{M_i\}$, leaving $k-i$ metarules in
$\mathcal{M}_1$. Thus, at step $i = 1$, $\mathcal{M}_1 = \mathcal{M}$ while at
step $i = k$, $\mathcal{M}_1 = \emptyset$. In each step $i$ we train TOIL-2 and
TOIL-3 with a set of matrix or punch metarules (described in the following
section), respectively, then we replace all the metarules in $\mathcal{M}_2$
with the output of TOIL-2 and replace all the metarules in $\mathcal{M}_3$ with
the output of TOIL-3 (in other words, we renew $\mathcal{M}_2$ and
$\mathcal{M}_3$ in each step). Then, in leg 1 we train Louise with the metarules
in $\mathcal{M}_1$ only; in leg 2 we train Louise with the metarules in
$\mathcal{M}_1 \cup \mathcal{M}_2$; and in leg 3 we train Louise with the
metarules in $\mathcal{M}_1 \cup \mathcal{M}_3$.  As to examples, at each step
$i$ we sample at random and without replacement 50\% of the examples in each of
$E^+$ and $E^-$ as a training partition and hold the rest out as a testing
partition. We sample a new pair of training and testing partitions in each leg
of each step of each run of the experiment and perform a learning attempt with
Louise on the training partition. We measure the accuracy of the hypothesis
learned in each learning attempt on the testing partition, and the duration of
the learning attempt in seconds. We measure accuracy and duration in two
separate learning attempts for each leg. In total we perform (10 runs *
$|\mathcal{M}|$ steps * 3 legs * 2 measurements) distinct learning attempts,
each with a new randomly chosen training and testing partition. We set a time
limit of 300 sec. for each learning attempt. If a learning attempt exhausts the
time limit we calculate the accuracy of the empty hypothesis on the testing
partition. Finally, we return the mean and standard error of the accuracy and
duration for the learning attempts at the same step of each leg over all 10
runs.

We run all experiments on a PC with 32 8-core Intel Xeon E5-2650 v2 CPUs clocked
at 2.60GHz with 251 Gb of RAM and running Ubuntu 16.04.7.

\subsection{Experiment datasets}
\label{subsec:Experiment datasets}

\begin{table}[t]
	\centering
	\begin{tabular}{lllll}
		\multicolumn{4}{l}{\textbf{Experiment datasets \& MIL problems}} \\
		\toprule
		& $\boldsymbol{|E^+|}$ & $\boldsymbol{|E^-|}$ & $\boldsymbol{|B|}$ & $\boldsymbol{|\mathcal{M}|}$ \\
		\midrule
		Grid world          & 81   & 0  & 16 & 14 \\
		Coloured graph (1)  & 108  & 74 & 9  & 14 \\
		M:tG Fragment       & 1348 & 0  & 60 & 14 \\
		\bottomrule
	\end{tabular}
\caption{Dataset summary. $|B|$: number of predicates defined in background
knowledge. $|\mathcal{M}|$: starting number of sort metarules.}
\label{tab:mil_problem_summary}
\end{table}

We reuse the datasets described in \cite{Patsantzis2021a}. These comprise: a)
Grid World, a grid-world generator for robot navigation problems; b) Coloured
Graph, a generator of fully-connected coloured graphs where the target predicate
is a representation of the connectedness relation and comprising four separate
datasets with different types of classification noise in the form of
misclassified examples (false positives, false negatives, both kinds and none);
and c) M:tG Fragment, a hand-crafted grammar of the Controlled Natural Language
of the Collectible Card Game, ``Magic: the Gathering" where examples are strings
entailed by the grammar. Table \ref{tab:mil_problem_summary} summarises the MIL
problem elements of the three datasets. We refer the reader to
\cite{Patsantzis2021a} for a full description of the three datasets.

Instead of the metarules defined in the experiment datasets we start each
experiment by initialising $\mathcal{M}$ to the set of 14 $H^2_2$ metarules in
\cite{Cropper2015}, listed in Table \ref{tab:canonical_h22_metarules}, which we
call the \emph{canonical $H^2_2$} set. We replace them with specialisations of
the matrix metarules \emph{Meta-dyadic} and \emph{Meta-monadic} from Table
\ref{tab:most_general_H22_metarules} and the punch metarules \emph{TOM-3} for
M:tG Fragment or \emph{TOM-2, TOM-3}, from Table \ref{tab:third_order_metarules}
otherwise. We limit over-generation in metarule specialisation as described in
Section \ref{sec:Implementation} by limiting metarule specialisation attempts to
1 for M:tG Fragment; and sub-sampling 50\% of $E^+$ at the start of each
metarule learning attempt, for Grid World and Coloured Graph.

Our configuration of the three experimental datasets is identical to that in
\cite{Patsantzis2021a} with the exception of the Grid World dataset, which we
configure to generate a grid world of dimensions $3 \times 3$. The resulting
learning problem is trivial, but hypotheses learned with metarules derived by
TOIL for worlds of larger dimensions tend to be extremely large (hypothesis
cardinalities upwards of 6,000 clauses are logged in preliminary experiments),
consuming an inordinate amount of resources during \emph{evaluation}. By
comparison, \cite{Patsantzis2021a} report a hypothesis of 2,567 clauses for a $5
\times 5$ world (as in our preliminary experiment). This observation indicates
that future work must address over-generation by TOIL. Still, the size of the
learned hypotheses serves as a stress test for our implementation.

\subsection{Experiment results}
\label{subsec:Experiment results}

\begin{figure}[t]
	\centering
	\subfloat[Grid World \label{fig:grid_world_acc}]{\includegraphics[width=0.33\columnwidth]{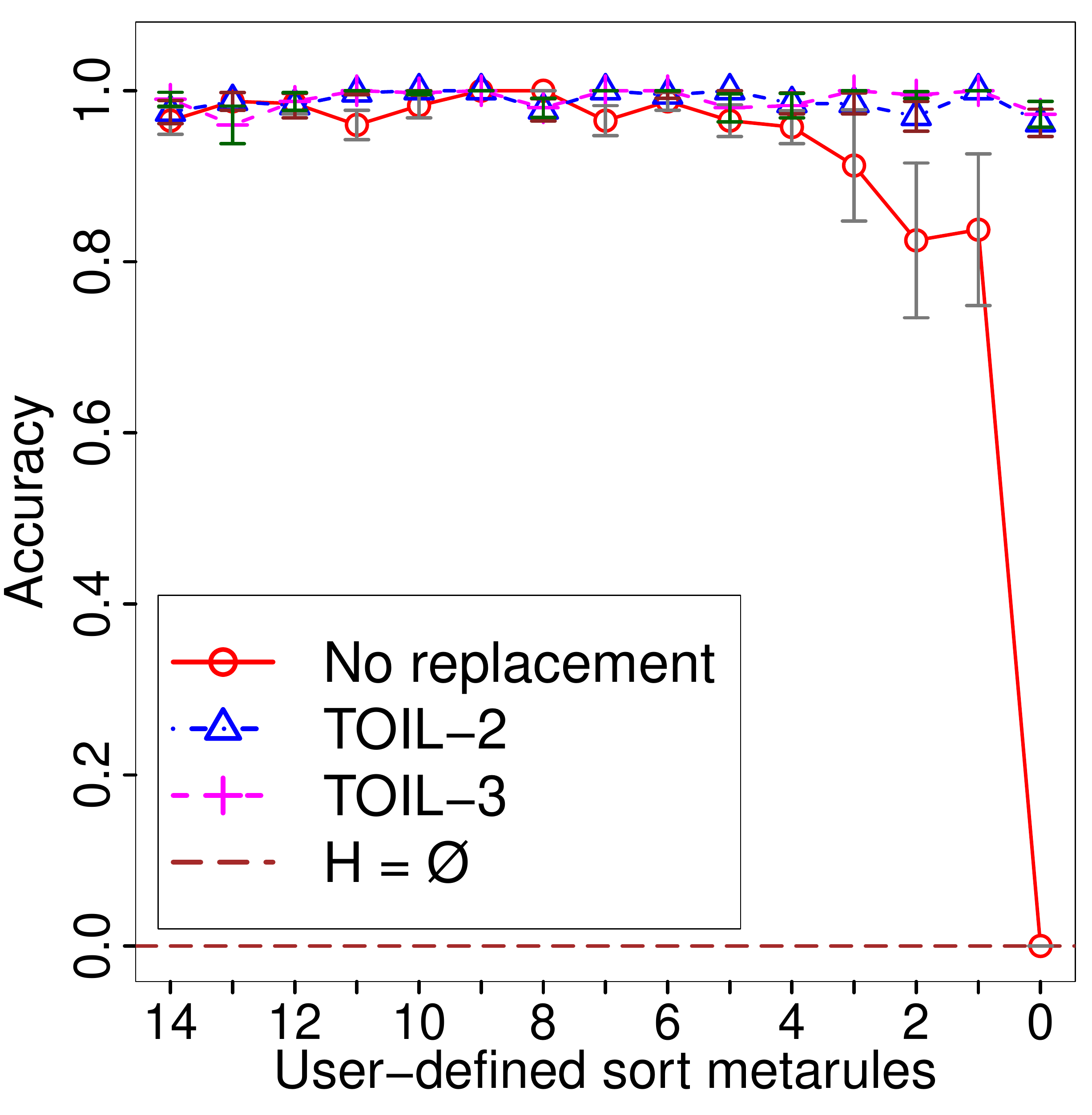}} 
	\subfloat[M:tG Fragment \label{fig:mtg_fragment_acc}]{\includegraphics[width=0.33\columnwidth]{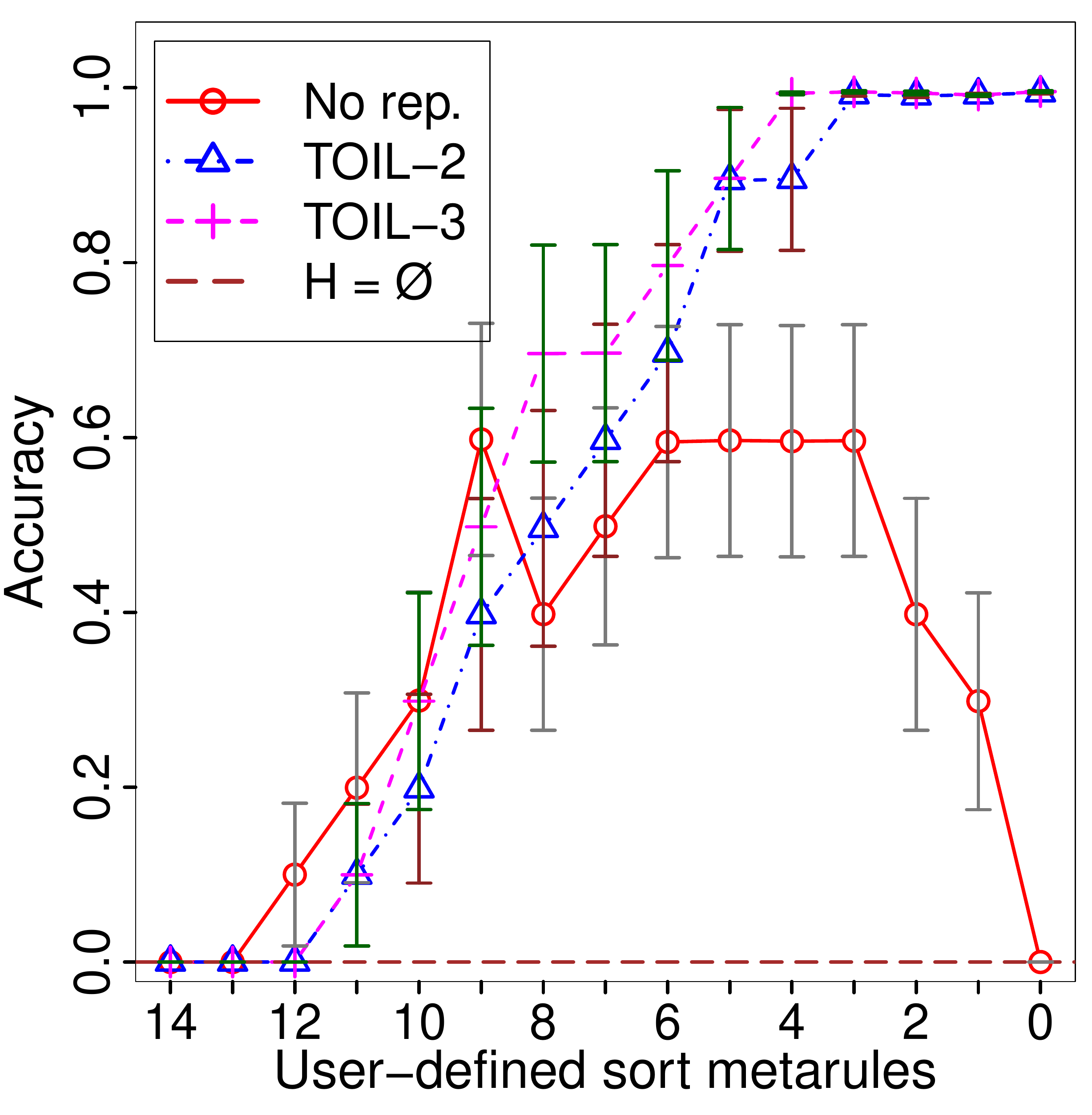}} 
	\subfloat[Coloured Graph - No Noise \label{fig:redundant_acc}]{\includegraphics[width=0.33\columnwidth]{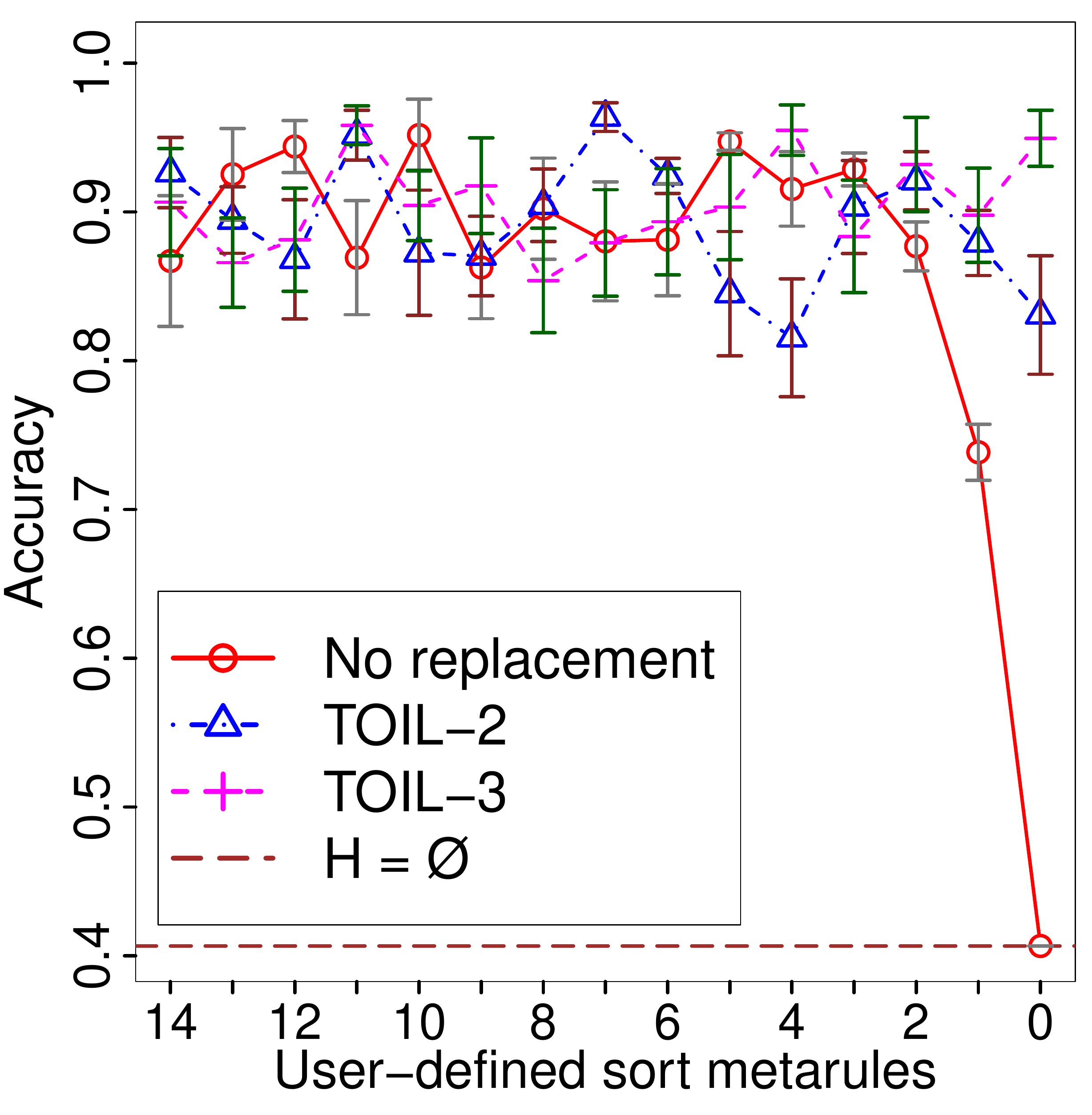}} \\
	\subfloat[Col. Graph - Ambiguities  \label{fig:ambiguities_acc}]{\includegraphics[width=0.33\columnwidth]{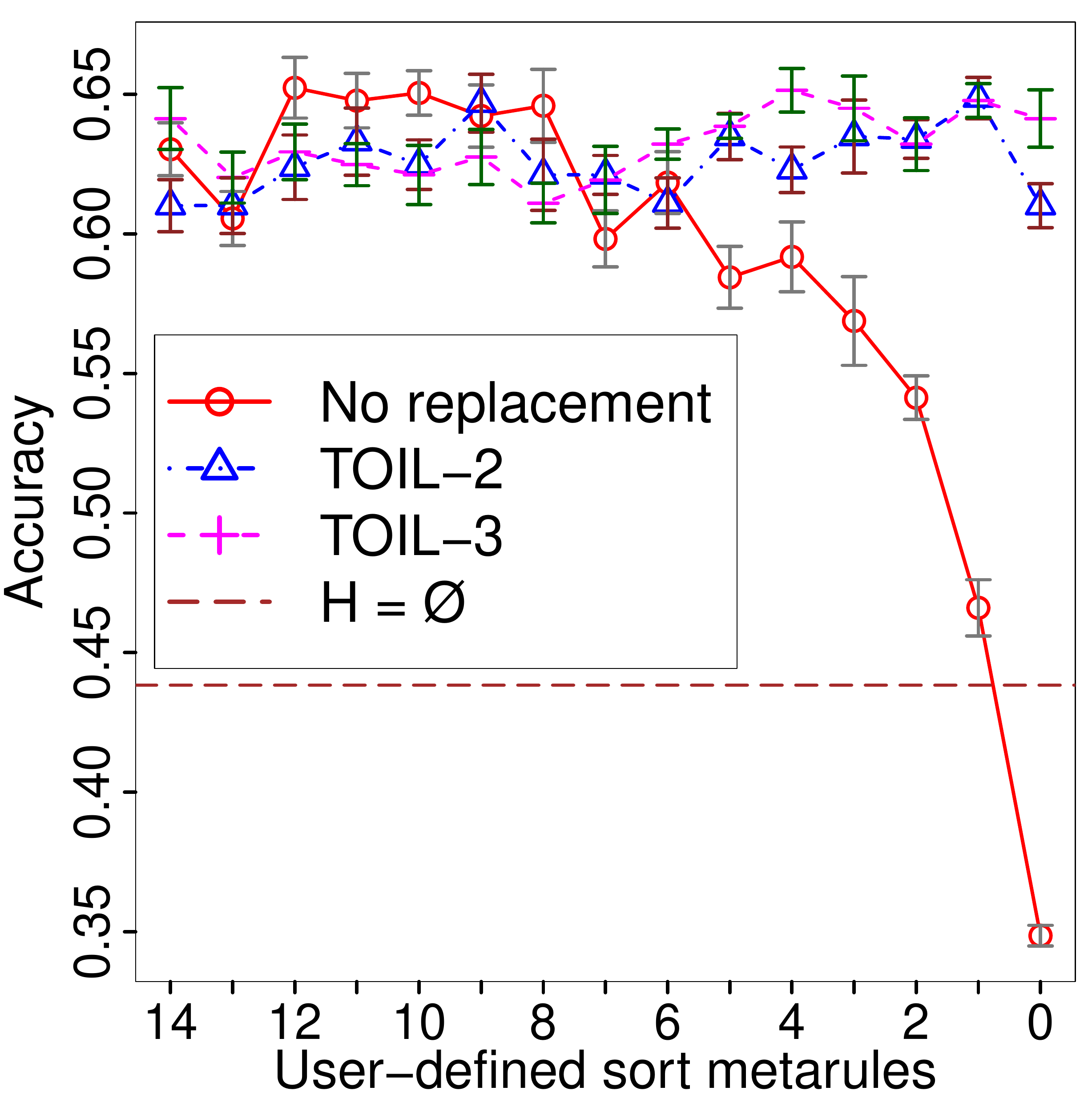}} 
	\subfloat[Col. Graph - False Positives \label{fig:false_positives_acc}]{\includegraphics[width=0.33\columnwidth]{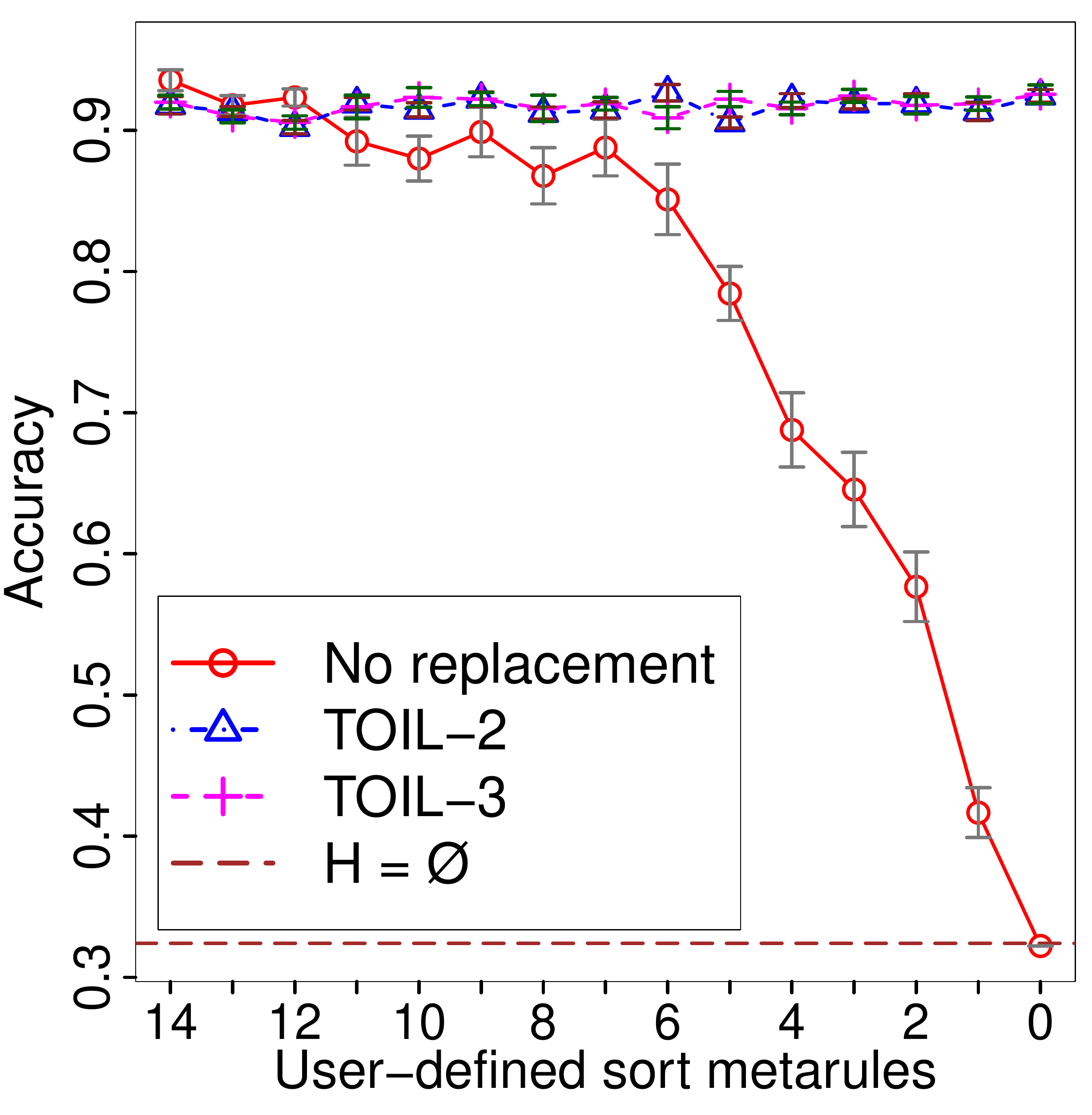}} 
	\subfloat[Col. G. - False Negatives \label{fig:false_negatives_acc}]{\includegraphics[width=0.33\columnwidth]{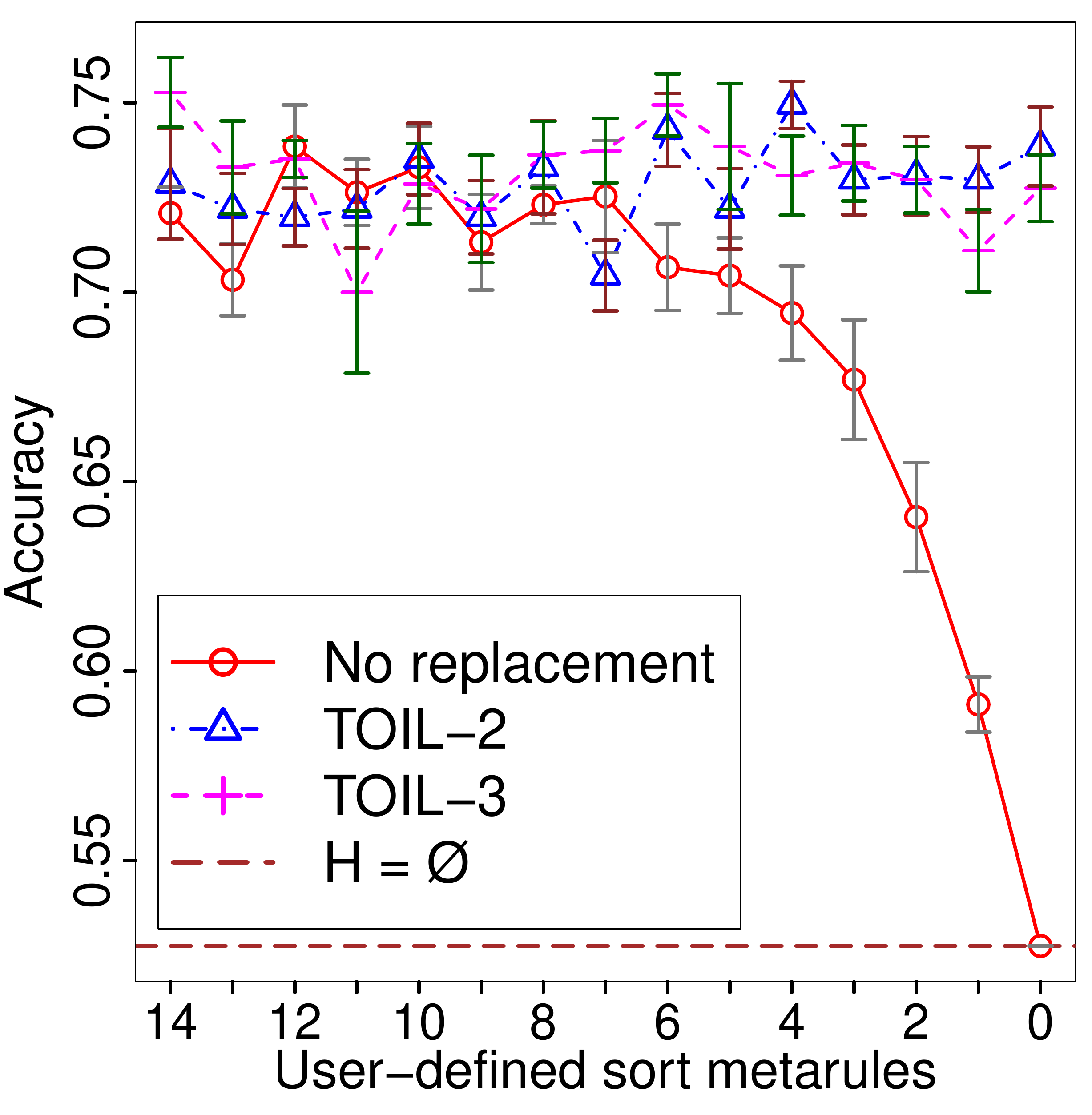}} 
\caption{
Experiment results measuring accuracy. User-defined sort metarules are
progressively reduced throughout the experiment. Red circles: no replacement.
Blue triangles: user-defined sort metarules are replaced by specialisations of
matrix metarules derived by TOIL-2. Magenta crosses: user-defined sort metarules
are replaced by specialisations of punch metarules derived by TOIL-3. Brown
dotted line: baseline (empty hypothesis). x-axis: number of remaining sort
metarules; y axis: accuracy on testing partition (50\% of $E^+,E^-$). Error
bars: standard error.
}
\label{fig:experiment_results_accuracy}
\end{figure}

\begin{figure}[t]
	\centering
	\subfloat[Grid World \label{fig:grid_world_time}]{\includegraphics[width=0.33\columnwidth]{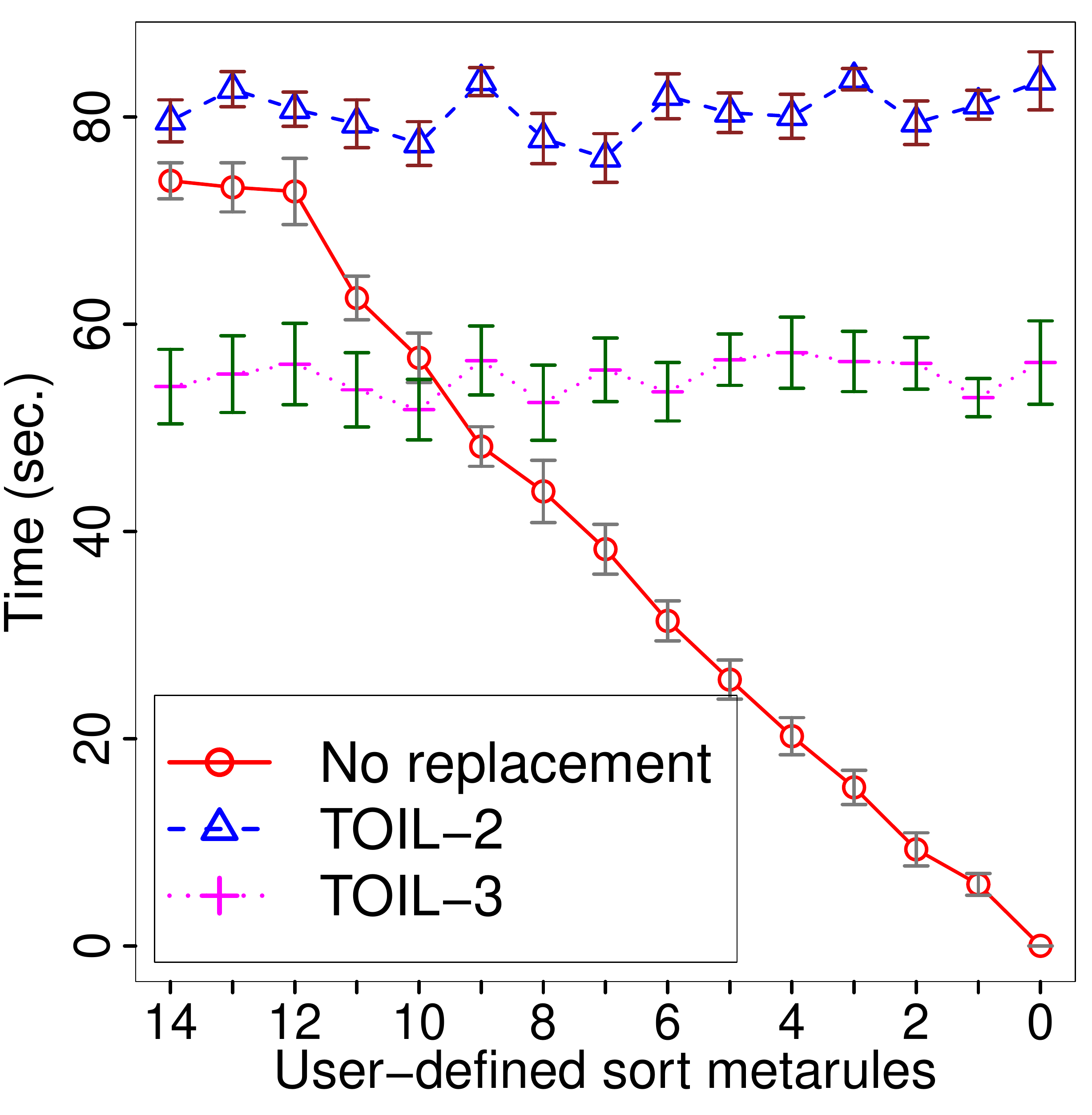}} 
	\subfloat[M:tG Fragment \label{fig:mtg_fragment_time}]{\includegraphics[width=0.33\columnwidth]{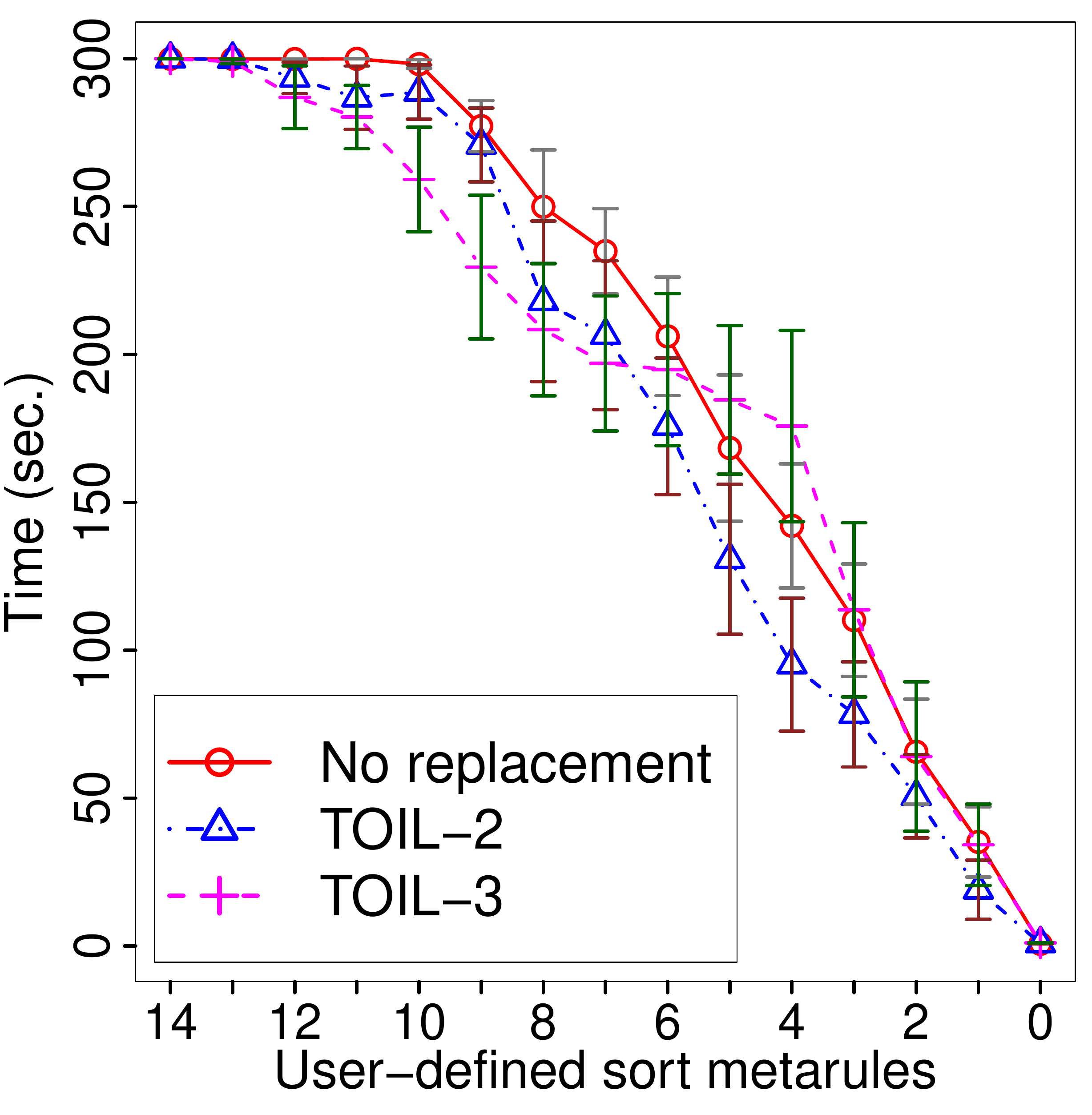}} 
	\subfloat[Coloured Graph \label{fig:redundant_time}]{\includegraphics[width=0.33\columnwidth]{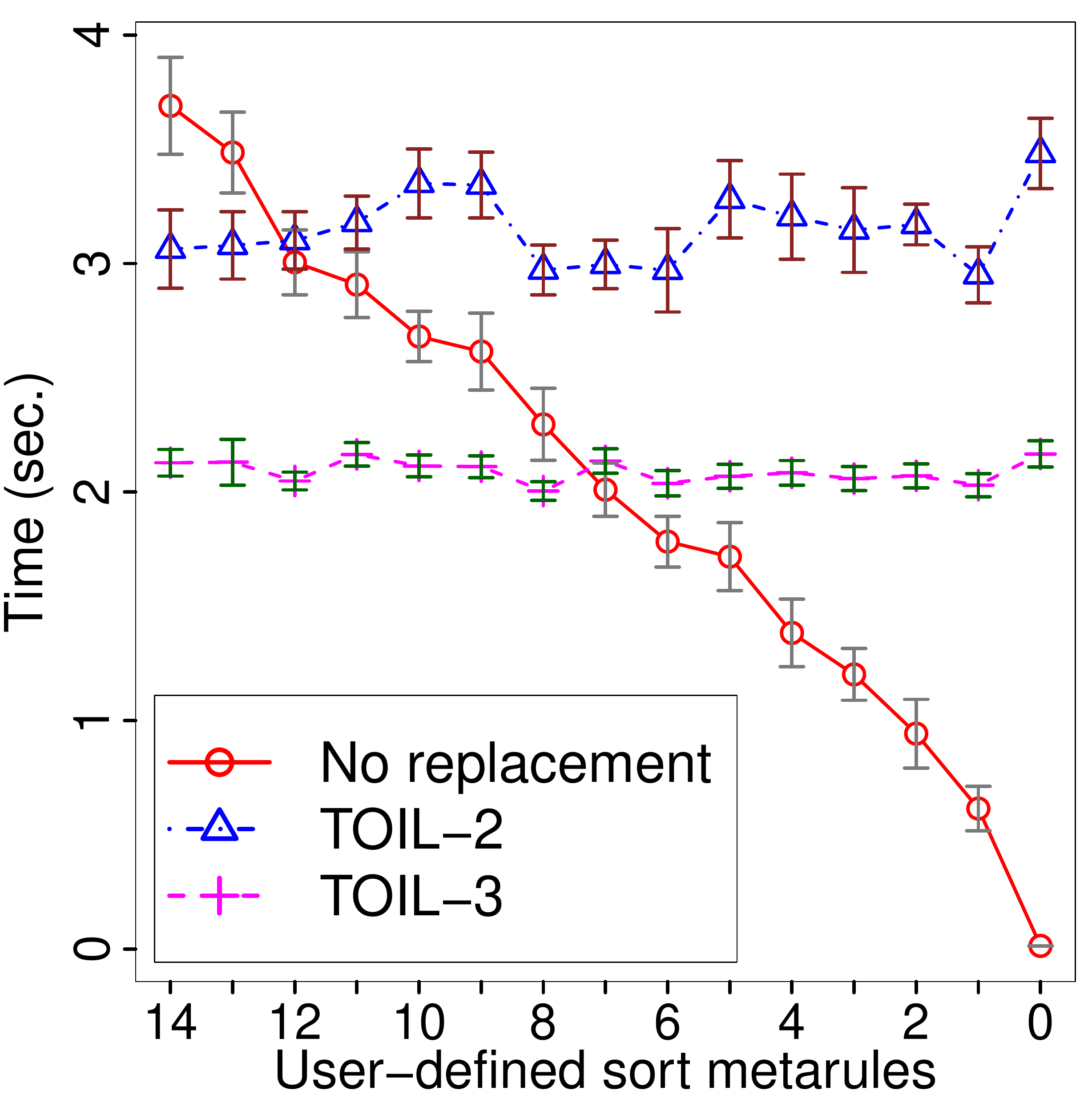}} \\
	\subfloat[Col. Graph - Ambiguities \label{fig:ambiguities_time}]{\includegraphics[width=0.33\columnwidth]{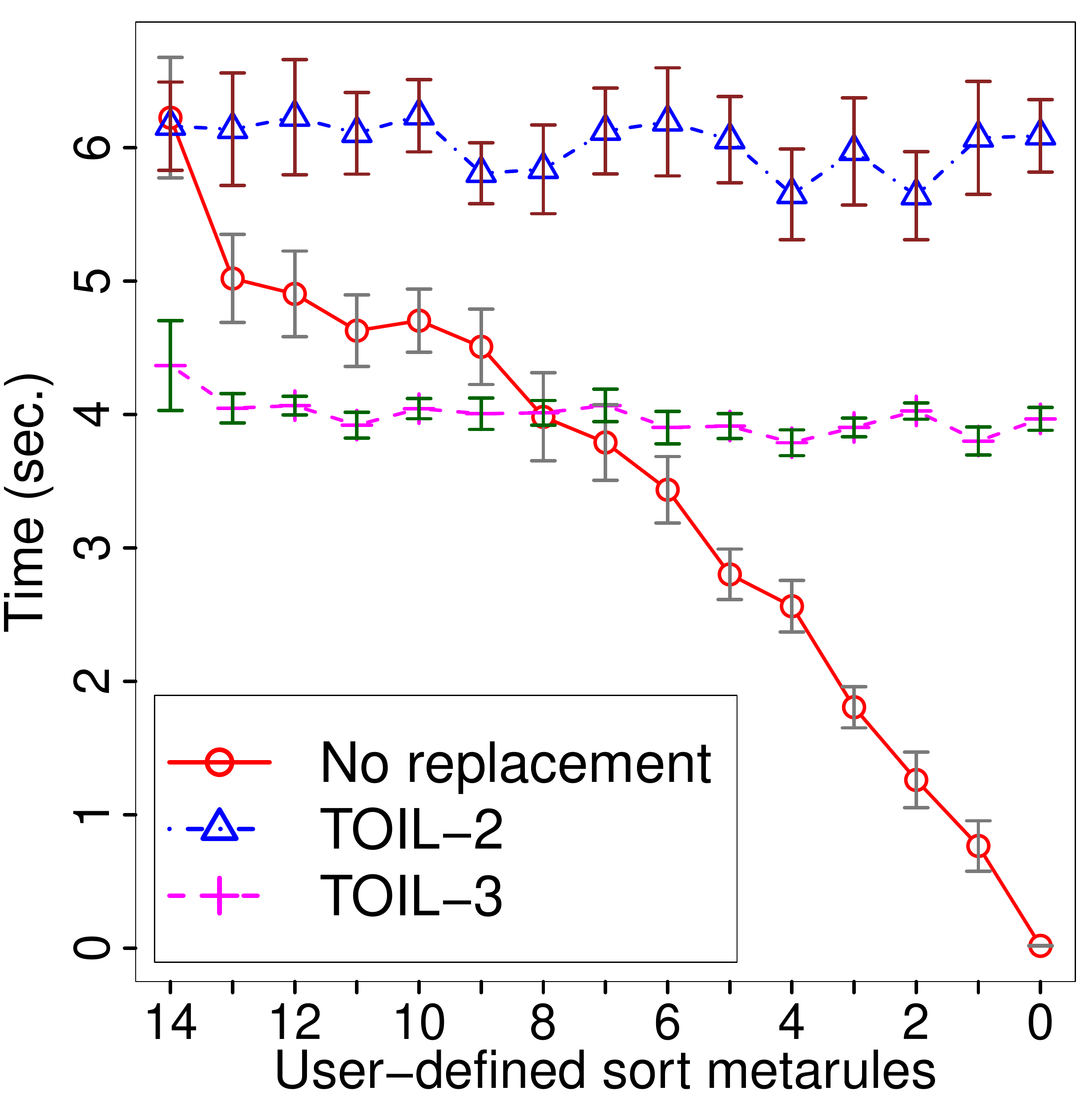}}
	\subfloat[Col. Graph - False Positives \label{fig:false_positives_time}]{\includegraphics[width=0.33\columnwidth]{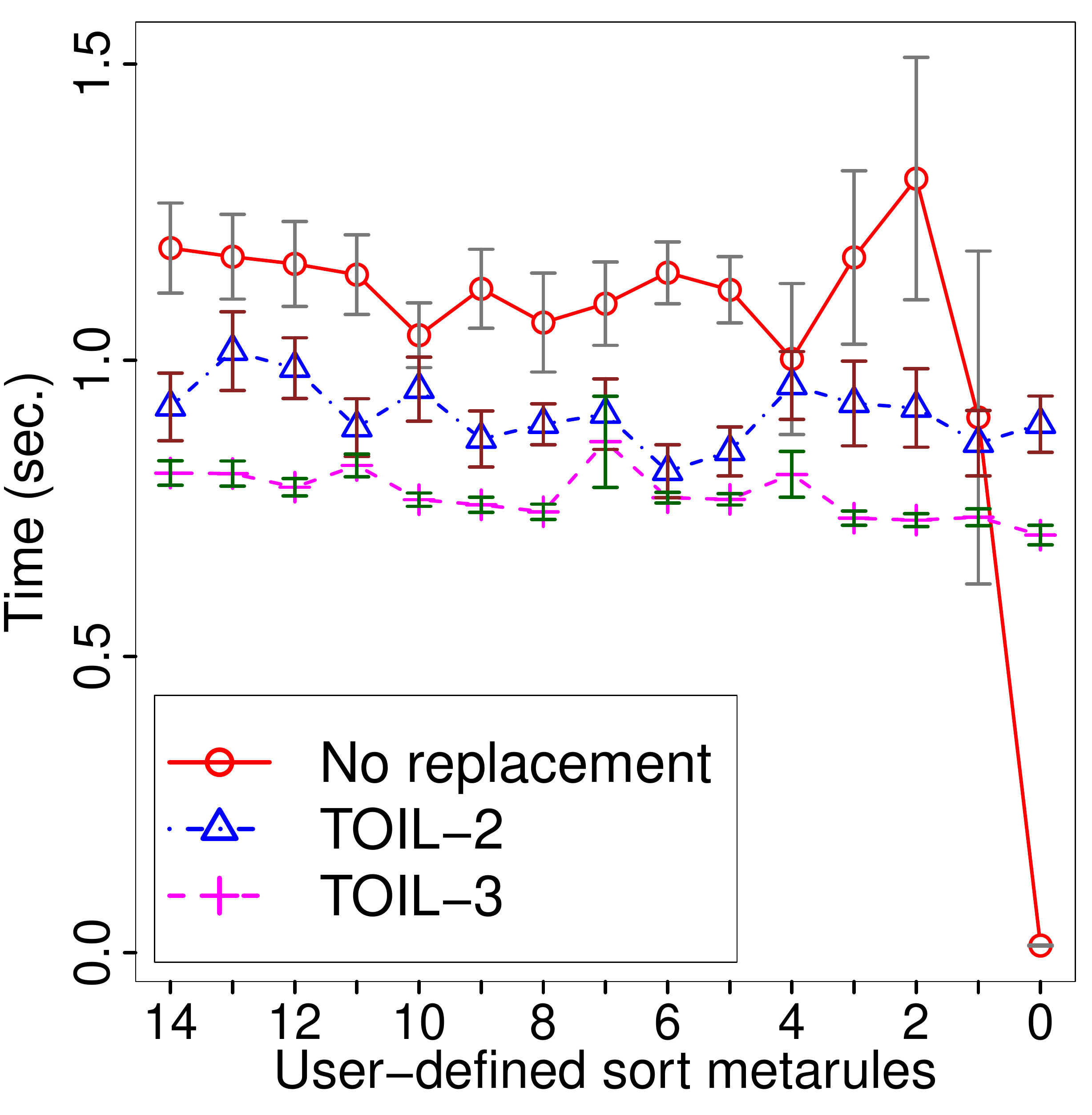}}
	\subfloat[Col. G. - False Negatives \label{fig:false_negatives_time}]{\includegraphics[width=0.33\columnwidth]{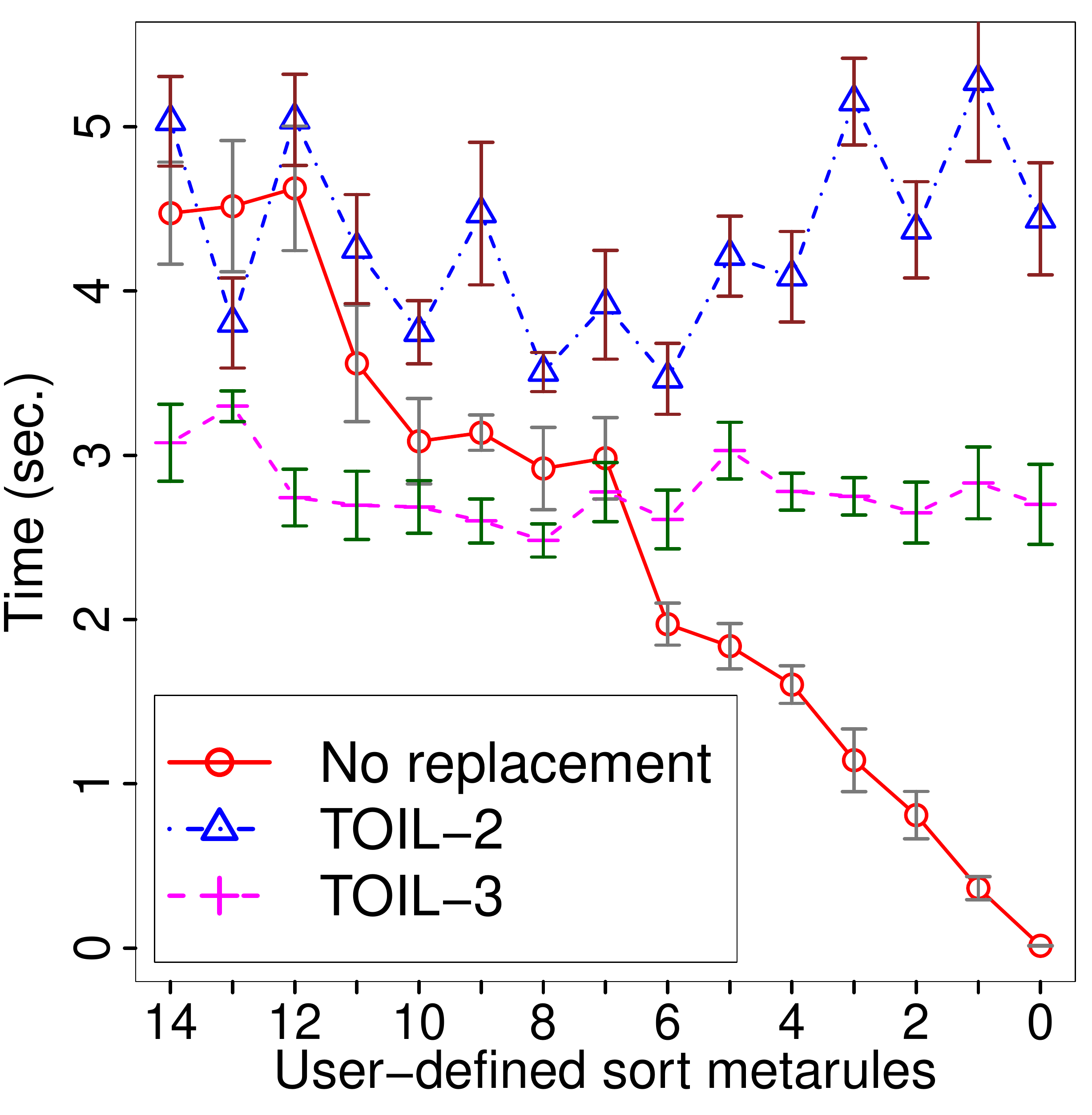}}
\caption{
Experiment results measuring running times. Lines, axes and error bars as in
Figure \ref{fig:experiment_results_accuracy}. No baseline for timing
experiments.
}
\label{fig:experiment_results_time}
\end{figure}

Figure \ref{fig:experiment_results_accuracy} lists the results of the
experiments measuring predictive accuracy. We immediately observe that in the
two legs of the experiment where user-defined metarules are replaced by learned
metarules, marked by ``TOIL-2" and ``TOIL-3", Louise's accuracy is maintained,
while it degrades in the leg of the experiment where metarules are reduced
without replacement, marked ``No replacement". These results support
Experimental Hypothesis \ref{hyp:predictive_accuracy}.

Figure \ref{fig:experiment_results_time} lists the results of the six
experiments measuring training times. We observe that training times for the
``No replacement" leg of the experiments decrease as the number of user-defined
metarules decreases, but remain more or less constant for the other two legs as
removed metarules are replaced by metarules learned by TOIL-2 and -3. In the
M:tG Fragment dataset, all but a single metarule in the canonical set, the
\emph{Chain} metarule, are redundant. TOIL-2 and -3 only learn the \emph{Chain}
metarule in their two legs of the experiment and so, as redundant metarules are
removed and not replaced, training times decrease in all three legs of that
experiment. These results support Experimental Hypothesis
\ref{hyp:running_time}.

\subsubsection{Learned metarules}
\label{subsubsec:Learned metarules}

\begin{table}[t]
	\centering
	\begin{tabular}{l}
		\textbf{TOIL-3 output for Coloured Graph - False Positives} \\
		\toprule
		$\% (Hom-1) \; \exists.P,Q \; \forall.x,y: P(x,y)\leftarrow Q(x,y)$ \\
		$\% (Hom-2) \; \exists.P,Q \; \forall.x,y: P(x,y)\leftarrow Q(y,x)$ \\
		$\% (Hom-3) \; \exists.P,Q,R \; \forall.x,y: P(x,y)\leftarrow Q(x,y),R(x,y)$ \\
		$\% (Hom-4) \; \exists.P,Q,R \; \forall.x,y: P(x,y)\leftarrow Q(x,y),R(y,x)$ \\
		$\% (Hom-5) \; \exists.P,Q,R \; \forall.x,y,z: P(x,y)\leftarrow Q(x,z),R(y,z)$ \\
		$\% (Hom-6) \; \exists.P,Q,R \; \forall.x,y,z: P(x,y)\leftarrow Q(x,z),R(z,y)$ \\
		$\% (Hom-7) \; \exists.P,Q,R \; \forall.x,y: P(x,y)\leftarrow Q(y,x),R(x,y)$ \\
		$\% (Hom-8) \; \exists.P,Q,R \; \forall.x,y: P(x,y)\leftarrow Q(y,x),R(y,x)$ \\
		$\% (Hom-9) \; \exists.P,Q,R \; \forall.x,y,z: P(x,y)\leftarrow Q(y,z),R(x,z)$ \\
		$\% (Hom-10) \; \exists.P,Q,R \; \forall.x,y,z: P(x,y)\leftarrow Q(y,z),R(z,x)$ \\
		$\% (Hom-11) \; \exists.P,Q,R \; \forall.x,y,z: P(x,y)\leftarrow Q(z,x),R(y,z)$ \\
		$\% (Hom-12) \; \exists.P,Q,R \; \forall.x,y,z: P(x,y)\leftarrow Q(z,x),R(z,y)$ \\
		$\% (Hom-13) \; \exists.P,Q,R \; \forall.x,y,z: P(x,y)\leftarrow Q(z,y),R(x,z)$ \\
		$\% (Hom-14) \; \exists.P,Q,R \; \forall.x,y,z: P(x,y)\leftarrow Q(z,y),R(z,x)$ \\
		\bottomrule
	\end{tabular}
\caption{Metarules learned by TOIL-3 for the Coloured Graph - False Positives
experiment collected from logging output during execution; these are exactly the
14 Canonical $H^2_2$ metarules in Table \ref{tab:canonical_h22_metarules}.}
\label{tab:learned_metarules}
\end{table}

During our experiments, the metarules learned by TOIL are logged to the command
line of the executing system. We could thus examine and will now discuss
examples of the metarules learned during execution.

For the M:tG Fragment dataset, we observed that TOIL-2 and -3 both learned a
single metarule, the \emph{Chain} metarule listed in Table
\ref{tab:canonical_h22_metarules}. The target theory for M:tG Fragment is a
grammar in Definite Clause Grammar form \citep{Colmerauer1978, Kowalski74},
where each clause is indeed an instance of \emph{Chain}. For this dataset, TOIL
was able to learn the set of metarules that would probably also be chosen by a
user.

For the Grid World dataset, TOIL-2 and -3 both learned a set of 22 $H^2_2$
metarules including the canonical set and 4 metarules with a single variable in
the head literal, e.g. $P(x,x)\leftarrow Q(x,y), R(x,y)$ or $P(x,x)\leftarrow
Q(x,y), R(y,x)$ that are useful to represent solutions of navigation tasks
beginning and ending in the same ``cell" of the grid world. Such metarules may
be seen as over-specialisations, but they are fully-connected sort metarules
which suggests that the constraints imposed on metarule specialisation to ensure
only fully-connected metarules are returned, described in Section
\ref{sec:Implementation}, are correctly defined.

Table \ref{tab:learned_metarules} lists metarules learned by TOIL-2 and -3 for
the Coloured Graph - False Positives dataset. The learned metarules are exactly
the set of 14 Canonical $H^2_2$ metarules in Table
\ref{tab:canonical_h22_metarules}. \cite{Cropper2015} show that the 14 canonical
$H^2_2$ metarules are reducible to a minimal set including only Inverse and
Chain, therefore returning the entire canonical set is redundant. This is an
example of the over-generation discussed in Section \ref{sec:Implementation},
that TOIL attempts to control by limiting the number of attempted metarule
specialisations. Logical minimisation by Plotkin's program reduction algorithm,
as described by \cite{Cropper2015}, could also be of help to reduce redundancy
in an already-learned set of metarules, although TOIL may be overwhelmed by
over-generation before reduction has a chance to be applied.  In any case,
over-generation is a clear weakness of our approach and must be further
addressed by future work.

\section{Conclusions and future work}
\label{sec:Conclusions and future work}

\subsection{Summary}
\label{subsec:Summary}

We have presented a novel approach for the automatic derivation of metarules for
MIL, by MIL. We have shown that the user-defined fully-connected second-order
\emph{sort metarules} used in the MIL literature can be derived by
specialisation of the most-general second-order \emph{matrix metarules} in a
language class, themselves derivable by specialisation of third-order
\emph{punch metarules} with literals that range over the set of second-order
literals. We have shown that metarule languages are enumerable in time
polynomial to the number of literals in punch metarules. We have defined two
methods of metarule specialisation, v- and l- specialisation and shown that
they are performed by MIL. We have proposed a modification of the MIL clause
construction operator to return fully connected second-order sort metarules,
rather than first-order clauses and proved its correctness. We have partially
implemented the modified MIL operator as TOIL, a new sub-system of the MIL
system Louise, and presented experiments demonstrating that metarules
automatically derived by TOIL can replace user-defined metarules while
maintaining predictive accuracy and training times.

\subsection{Future work}
\label{subsec:Future work}

\begin{table}[t]
	\centering
	\begin{tabular}{ll}
		%\textbf{Transferring learned generalisations} \\
		%\textbf{Transfer of learned metarules as analogy-making} \\
		%\textbf{Transferring learned metarules as analogies} \\
		%\textbf{Reusing learned metarules as analogy formation} \\
		\textbf{Forming analogies by transferring learned metarules} \\
		%\textbf{Forming analogies by transferring learned, non-}$\boldsymbol{H^2_2}$ \textbf{ metarules} \\
		\toprule
		$E^+_1 = \, \{ parents(kostas,dora,stassa) \}$, $E^-_1 = \emptyset$\\
		$B_1 = \, \{ father(kostas,stassa), mother(dora,stassa) \}$ \\
		$\mathcal{M}_1 = \, \{ (TOM-3) \; P \leftarrow Q, R$ \}\\
		\textsc{TOIL-3}$(E^+_1,E^-_1,B_1,\mathcal{M}_1) = $ $\, \{(M_1) \; P(x,y,z) \leftarrow Q(x,z), R(y,z) \}$\\
		\textsc{Louise}$(E^+_1,E^-_1,B_1,\{M_1\}) = $ $\, \{parents(x,y,z) \leftarrow father(x,z), mother(y,z) \}$ \\
		\midrule
		$E^+_2 = \, \{ bounded\_by(1,2,3), bounded\_by(3,2,1) \}$, $E^-_2 = \emptyset$\\
		$B_2 = \, \{ lt(1,3), lt(1,2), lt(2,3), gt(2,1), gt(3,1), gt(3,2) \}$ \\
		\textsc{Louise}$(E^+_2,E^-_2,B_2,\{M_1\}) = $ $\, \{bounded\_by(x,y,z) \leftarrow lt(x,z), lt(y,z)$ \\
		\qquad \qquad \qquad \qquad \qquad \qquad \quad \quad $bounded\_by(x,y,z) \leftarrow gt(x,z), gt(y,z)$ \}\\
		\bottomrule
	\end{tabular}
\caption{
This table illustrates the use of metarules as a data structure for transferring
generalisations between problems. TOIL-3 was used to learn a fully-connected
sort metarule, $M_1$, from examples and background knowledge of $parent/3$, in
$E^+_1,E^-_1,B_1$. Louise then learned a definition of $parent/3$ from $E^+_1,
E^-_1, B_1$ and $M_1$. Later, Louise learned a definition of $bounded\_by/3$
from $E^+_2,E^-_2,B_2$ and $M_1$. Thus, $M_1$ was used to form an analogy
between $parent/3$ and $bounded\_by/3$. Note that $M_1$ is outside $H^2_2$.
}
\label{tab:analogies}
\end{table}

The major practical limitations of our approach are the incomplete state of its
implementation and its over-generation of metarule specialisations.

Our prototype implementation of Algorithms \ref{alg:vl_specialisation} and
\ref{alg:lift} in TOIL is only capable of v-specialisation. Work is under way to
complete the implementation with the capability for l-specialisation. In Section
\ref{sec:Implementation}, we have left a formal treatment of TOIL to the time
this work is complete. Further work would improve the look-ahead heuristic in
Algorithm \ref{alg:look_ahead} and our ability to limit attempted metarule
specialisations to reduce over-generation. In general, we do not know of a good,
principled (as in non-heuristic) and efficient approach to derive just enough
metarules to solve a problem, without deriving too many and over-generalising.

Conversely to over-generation, TOIL also exhibits a tendency to produce
metarules that are over-\emph{specialised} to the examples in a MIL problem, a
form of over-fitting. This seems to be a limitation of TOIL's look-ahead
heuristic listed in Algorithm \ref{alg:look_ahead} and used to ensure learned
metarules are fully-connected. Future work should look for a principled approach
to replace this heuristic, also.

TOIL-3 is capable of learning new metarules with literals of arbitrary arities,
as illustrated in table \ref{tab:analogies}. We haven't demonstrated this
important ability with experiments. Additionally, we have not presented any
empirical results measuring training times for TOIL itself---only for Louise.
Theoretical results in Section \ref{subsec:Cardinality of metarule languages}
predict that learning metarules should be time-consuming, especially for larger
metarule languages, and we have observed this while executing the experiments in
Section \ref{sec:Experiments}, although more so for TOIL-3 than TOIL-2.

Our theoretical framework described in Section \ref{sec:Framework} extends
$\theta$-subsumption to metarules. It remains to be seen if the related
frameworks of relative subsumption and relative entailment
\citep{Nienhuys-Cheng1997} can also be extended to metarules. v- and
l-specialisation seem to be related to Shapiro's refinement operators
\citep{Shapiro1983}, a point also made about metarules in general by
\cite{Cropper2015} but we haven't explored this relation in this work.

Corollary \ref{cor:minimal_sets} suggests that classes of learning problems can
be solved by the same sets of metarules, as long as suitable solutions belong to
the same metarule language. This observation introduces the possibility of
transferring generalisations, in the form of learned metarules, across learning
problems or problem domains thus in a sense forming analogies, a capability
poorly represented in modern machine learning---and, in general, AI\-/systems
\citep{MitchellMellanie2021b}. Such a capability would however rely on a method
to determine the relevance of metarules to a problem; currently, no such method
is known. Table \ref{tab:analogies} illustrates the transfer of learned
metarules as analogies between problems.

Future work should test the accuracy of the metarules learned by TOIL with other
systems that use metarules, besides Louise, for example Metagol \citep{metagol},
Popper \citep{Cropper2021} and ALPS \citep{Xujie2019}.

\section{Acknowledgements}
\label{sec:Acknowledgements}

The first author acknowledges support from the UK's EPSRC for financial support
of her studentship. The second author acknowledges support from the UK's EPSRC
Human-Like Computing Network, for which he acts as director. We thank the
anonymous reviewers for their diligent and knowledgeable reviews that have
helped us significantly improve our work.

\section{Declarations}
\label{Declarations}

Author 1 wrote all sections of the paper. Author 2 provided feedback and
corrections on all sections of the paper. The authors have no conflicts of
interest to disclose. Ethics approval, consent to participate and consent for
publication were not required. Code and data have been made available in Section
\ref{sec:Experiments}.

\bibliographystyle{spbasic}
\bibliography{mybib}

\begin{thebibliography}{32}
\providecommand{\natexlab}[1]{#1}
\providecommand{\url}[1]{{#1}}
\providecommand{\urlprefix}{URL }
\expandafter\ifx\csname urlstyle\endcsname\relax
  \providecommand{\doi}[1]{DOI~\discretionary{}{}{}#1}\else
  \providecommand{\doi}{DOI~\discretionary{}{}{}\begingroup
  \urlstyle{rm}\Url}\fi
\providecommand{\eprint}[2][]{\url{#2}}

\bibitem[{{Ceri} et~al.(1989){Ceri}, {Gottlob}, and {Tanca}}]{Ceri1989}
{Ceri} S, {Gottlob} G, {Tanca} L (1989) What you always wanted to know about
  datalog (and never dared to ask). IEEE Transactions on Knowledge and Data
  Engineering 1(1):146--166

\bibitem[{Colmerauer(1978)}]{Colmerauer1978}
Colmerauer A (1978) Metamorphosis grammars, Springer Berlin Heidelberg, Berlin,
  Heidelberg, pp 133--188. \urlprefix\url{https://doi.org/10.1007/BFb0031371}

\bibitem[{Cormen et~al.(2001)Cormen, Leiserson, Rivest, and Stein}]{Cormen2001}
Cormen T, Leiserson C, Rivest R, Stein C (2001) Introduction to algorithms,
  second edition

\bibitem[{Cropper and Morel(2021)}]{Cropper2021}
Cropper A, Morel R (2021) Learning programs by learning from failures. Machine
  Learning \urlprefix\url{https://doi.org/10.1007/s10994-020-05934-z}

\bibitem[{Cropper and Muggleton(2016{\natexlab{a}})}]{Cropper2016}
Cropper A, Muggleton S (2016{\natexlab{a}}) Learning higher-order logic
  programs through abstraction and invention. In: Proceedings of the 25th
  International Joint Conference Artificial Intelligence (IJCAI 2016), IJCAI,
  pp 1418--1424,
  \urlprefix\url{http://www.doc.ic.ac.uk/\~shm/Papers/metafunc.pdf}

\bibitem[{Cropper and Muggleton(2015)}]{Cropper2015}
Cropper A, Muggleton SH (2015) {Logical minimisation of meta-rules within
  Meta-Interpretive Learning}. In: Proceedings of the 24th International
  Conference on Inductive Logic Programming, pp 65--78

\bibitem[{Cropper and Muggleton(2016{\natexlab{b}})}]{metagol}
Cropper A, Muggleton SH (2016{\natexlab{b}}) Metagol system.
  \urlprefix\url{https://github.com/metagol/metagol}

\bibitem[{Cropper and Tourret(2018)}]{Cropper2018}
Cropper A, Tourret S (2018) Derivation reduction of metarules in
  meta-interpretive learning. In: Riguzzi F, Bellodi E, Zese R (eds) Inductive
  Logic Programming, Springer International Publishing, Cham, pp 1--21

\bibitem[{Emde(1987)}]{Emde87}
Emde W (1987) Non-cumulative learning in metaxa.3. In: Proceedings of IJCAI-87,
  Morgan Kaufmann, pp 208--210

\bibitem[{Emde et~al.(1983)Emde, Habel, rainer Rollinger, Berlin, Kit, and
  Fr}]{Emde83}
Emde W, Habel CU, rainer Rollinger C, Berlin TU, Kit P, Fr S (1983) The
  discovery of the equator or concept driven learning. In: Proceedings of the
  8th International Joint Conference on Artificial Intelligence, Morgan
  Kaufmann, pp 455--458

\bibitem[{Evans and Grefenstette(2018)}]{Evans2018}
Evans R, Grefenstette E (2018) Learning explanatory rules from noisy data.
  Journal of Artificial Intelligence Research 61:1–64,
  \urlprefix\url{http://dx.doi.org/10.1613/jair.5714}

\bibitem[{Kaminski et~al.(2018)Kaminski, Eiter, and
  Inoue}]{Kaminski2018ExploitingAS}
Kaminski T, Eiter T, Inoue K (2018) Exploiting answer set programming with
  external sources for meta-interpretive learning. TPLP 18:571--588

\bibitem[{Kietz and Wrobel(1992)}]{Kietz92}
Kietz JU, Wrobel S (1992) Controlling the complexity of learning in logic
  through syntactic and task-oriented models. In: Inductive Logic Programming,
  Academic Press, pp 335--359

\bibitem[{Kowalski(1974)}]{Kowalski74}
Kowalski R (1974) Logic for problem solving. Memo No 75, March 1974, Department
  of Computational Logic, School of Artificial Intelligence, University of
  Edinburgh, \urlprefix\url{http://www.doc.ic.ac.uk/~rak/papers/Memo75.pdf}

\bibitem[{Lin et~al.(2014)Lin, Dechter, Ellis, Tenenbaum, Muggleton, and
  Dwight}]{Lin2014}
Lin D, Dechter E, Ellis K, Tenenbaum J, Muggleton S, Dwight M (2014) {Bias
  reformulation for one-shot function induction}. In: Proceedings of the 23rd
  European Conference on Artificial Intelligence, pp 525--530,
  \doi{10.3233/978-1-61499-419-0-525}

\bibitem[{Mitchell(2021)}]{MitchellMellanie2021b}
Mitchell M (2021) Abstraction and analogy-making in artificial intelligence.
  arXiv:210210717v1 [csAI] \urlprefix\url{https://arxiv.org/abs/2102.10717},
  \eprint{2102.10717}

\bibitem[{Morik(1993)}]{Morik1993}
Morik K (1993) Balanced Cooperative Modeling, Springer US, Boston, MA, pp
  109--127. \urlprefix\url{https://doi.org/10.1007/978-1-4615-3202-6\_6}

\bibitem[{Muggleton and Lin(2015)}]{Muggleton2015}
Muggleton S, Lin D (2015) {Meta-Interpretive Learning of Higher-Order Dyadic
  Datalog : Predicate Invention Revisited}. Machine Learning 100(1):49--73

\bibitem[{Muggleton and de~Raedt(1994)}]{Muggleton1994}
Muggleton S, de~Raedt L (1994) {Inductive Logic Programming: Theory and
  methods}. The Journal of Logic Programming 19-20(SUPPL. 1):629--679,
  \urlprefix\url{https://doi.org/10.1016/0743-1066(94)90035-3}

\bibitem[{Muggleton et~al.(2014)Muggleton, Lin, Pahlavi, and
  Tamaddoni-Nezhad}]{Muggleton2014}
Muggleton SH, Lin D, Pahlavi N, Tamaddoni-Nezhad A (2014) {Meta-interpretive
  learning: Application to grammatical inference}. Machine Learning
  94(1):25--49, \urlprefix\url{https://doi.org/10.1007/s10994-013-5358-3}

\bibitem[{Nienhuys-Cheng and de~Wolf(1997)}]{Nienhuys-Cheng1997}
Nienhuys-Cheng SH, de~Wolf R (1997) {Foundations of Inductive Logic
  programming}. Springer-Verlag, Berlin

\bibitem[{Patsantzis and Muggleton(2019{\natexlab{a}})}]{Louise}
Patsantzis S, Muggleton SH (2019{\natexlab{a}}) Louise system.
  \urlprefix\url{https://github.com/stassa/louise}

\bibitem[{Patsantzis and Muggleton(2019{\natexlab{b}})}]{Thelma}
Patsantzis S, Muggleton SH (2019{\natexlab{b}}) Thelma system.
  \urlprefix\url{https://github.com/stassa/thelma}

\bibitem[{Patsantzis and Muggleton(2021)}]{Patsantzis2021a}
Patsantzis S, Muggleton SH (2021) Top program construction and reduction for
  polynomial time meta-interpretive learning. Machine Learning
  \urlprefix\url{https://doi.org/10.1007/s10994-020-05945-w}

\bibitem[{Plotkin(1972)}]{Plotkin1972}
Plotkin G (1972) {Automatic Methods of Inductive Inference}. PhD thesis, The
  University of Edinburgh

\bibitem[{Robinson(1965)}]{Robinson1965}
Robinson JA (1965) A machine-oriented logic based on the resolution principle.
  J ACM 12(1):23–41, \urlprefix\url{https://doi.org/10.1145/321250.321253}

\bibitem[{Rouveirol(1994)}]{Rouveirol1994}
Rouveirol C (1994) Flattening and saturation: Two representation changes for
  generalization. Machine Learning 14(2):219--232,
  \doi{10.1023/A:1022678217288},
  \urlprefix\url{https://doi.org/10.1023/A:1022678217288}

\bibitem[{Shapiro(2004)}]{Shapiro1983}
Shapiro EY (2004) {Algorithmic Program Debugging}. The MIT Press,
  \urlprefix\url{https://doi.org/10.7551/mitpress/1192.001.0001}

\bibitem[{Si et~al.(2018)Si, Lee, Zhang, Albarghouthi, Koutris, and
  Naik}]{Xujie2018}
Si X, Lee W, Zhang R, Albarghouthi A, Koutris P, Naik M (2018) Syntax-guided
  synthesis of datalog programs. In: Proceedings of the 2018 26th ACM Joint
  Meeting on European Software Engineering Conference and Symposium on the
  Foundations of Software Engineering, Association for Computing Machinery, New
  York, NY, USA, ESEC/FSE 2018, p 515–527,
  \urlprefix\url{https://doi.org/10.1145/3236024.3236034}

\bibitem[{Si et~al.(2019)Si, Raghothaman, Heo, and Naik}]{Xujie2019}
Si X, Raghothaman M, Heo K, Naik M (2019) Synthesizing datalog programs using
  numerical relaxation. In: Proceedings of the Twenty-Eighth International
  Joint Conference on Artificial Intelligence, {IJCAI-19}, International Joint
  Conferences on Artificial Intelligence Organization, pp 6117--6124,
  \urlprefix\url{https://doi.org/10.24963/ijcai.2019/847}

\bibitem[{Stanley(2011)}]{Stanley1997}
Stanley RP (2011) Enumerative Combinatorics, Volume 1, 2nd Edition. Cambridge
  University Press

\bibitem[{Wrobel(1988)}]{Wrobel1988}
Wrobel S (1988) Design goals for sloppy modeling systems. International Journal
  of Man-Machine Studies 29(4):461 -- 477,
  \urlprefix\url{https://doi.org/10.1016/S0020-7373(88)80006-3}

\end{thebibliography}

\newpage
\section*{Appendix A Predicate invention in MIL: full description}
\label{app:Predicate invention in MIL: full description}

\begin{algorithm} [t]
	\caption{Resolution-based MIL clause construction}
	\label{alg:mil_full}
	\textbf{Input}: 1st- or 2nd- order literal $e$; $B^*$, $\mathcal{M}, I$, elements of a MIL problem; $\vartheta \Theta = \emptyset$.\\
	\textbf{Output}: $M\Theta$, a first-order instance of metarule $M \in \mathcal{M}$.
	\begin{algorithmic}[1]
		\Procedure{Construct}{$\neg e,B^*,\mathcal{M}, I, \vartheta \Theta$}
			\State Select $M \in \mathcal{M}$
			\If{$\exists \varrho P \supseteq \vartheta \Theta: head(M \varrho P) = e$}
				\State Set $\vartheta \Theta \Leftarrow \varrho P$
				\For{$l_i \in body(M \vartheta \Theta)$}
					\If{$\exists \sigma \Sigma \supseteq \vartheta \Theta: \{\neg l_i \sigma \Sigma\} \cup B^* \cup \mathcal{M} \vdash_{SLD} \square$} 
						\State Set $\vartheta \Theta \Leftarrow \sigma \Sigma$
					\Else
						\State Let $l_i = P(v_1, ..., v_n)$
						\State Set $\Omega \Leftarrow \{P/Q\}: \exists_{\in I} Q$
						\State Set $B^* \Leftarrow B^* \; \cup$ \textsc{Construct}$(\neg l_i \Omega, B^*,\mathcal{M}, I, \vartheta \Theta)$
					\EndIf
				\EndFor
			\State \textbf{Return} $M \Theta$
			\EndIf
		\State \textbf{Return} $\emptyset$
		\EndProcedure
	\end{algorithmic}
\end{algorithm}

\begin{algorithm} [t]
	\caption{Resolution-based MIL vl-specialisation}
	\label{alg:vl_specialisation_full}
	\textbf{Input}: 1st- or higher-order literal $e$; $B^*, I$ as in Alg. \ref{alg:mil_full}; punch or matrix metarules $\mathcal{M}$; $\vartheta \Theta = \emptyset$.\\
	\textbf{Output}: $\dot M$, a fully connected sort metarule.
	\begin{algorithmic}[1]
		\Procedure{VL-Specialise}{$\neg e,B^*,\mathcal{M}, I, \vartheta \Theta$}
			\State Select $M \in \mathcal{M}$
			\If{$\exists \varrho P \supseteq \vartheta \Theta: head(M \varrho P) = e$}
				\State Set $\vartheta \Theta \Leftarrow \varrho P$
				\For{$l_i \in body(M \vartheta \Theta)$}
					\If{$\exists \sigma \Sigma \supseteq \vartheta \Theta: \{\neg l_i \sigma \Sigma\} \cup B^* \vdash_{SLD} \square$}
						\State Set $\vartheta \Theta \Leftarrow \sigma \Sigma$
					\Else
						\State Let $l_i = P(v_1, ..., v_n)$
						\State Set $\Omega \Leftarrow \{P/Q\}: \exists_{\in I} Q$
						\State Set $\mathcal{M} \Leftarrow \mathcal{M} \; \cup$ \textsc{VL-Specialise}$(\neg l_i \Omega, B^*,\mathcal{M}, I, \vartheta \Theta)$
					\EndIf
				\EndFor
				\If{$M \vartheta \Theta$ is fully-connected}
					\State \textbf{Return} $M$.\textsc{Lift}$(\vartheta \Theta)$
				\EndIf
			\EndIf
		\State \textbf{Return} $\emptyset$
		\EndProcedure
	\end{algorithmic}
\end{algorithm}

In Section \ref{subsec:MIL as metarule specialisation} we have given a
simplified description of Algorithm \ref{alg:mil} omitting the recursive
resolution step that takes place during predicate invention. We have done this
to simplify the description of the algorithm and to isolate the specialisation
operation that is the primary subject of Section \ref{subsec:MIL as metarule
specialisation}. Algorithm \ref{alg:mil} is accurate as long as predicate
invention is not required. In this Appendix, Algorithm \ref{alg:mil_full} is a
more complete description of Algorithm \ref{alg:mil} that includes recursion and
predicate invention. Similarly, Algorithm \ref{alg:vl_specialisation_full} is a
more complete description, including the predicate invention step, of Algorithm
\ref{alg:vl_specialisation}. In our implementation of TOIL the propagation of
meta/substitution $\vartheta \Theta$ in line 11 of algorithms
\ref{alg:mil_full}, \ref{alg:vl_specialisation_full} is handled by the Prolog
engine.

\section*{Appendix B An example of metarule specialisation}
\label{app:Appendix B}

\begin{table}
	\centering
	\begin{tabular}{lll}
		\multicolumn{3}{l}{\textbf{Learning a grammar of the } $\boldsymbol{a^nb^n}$ \textbf{ CFG with TOIL and Louise}} \\
		\toprule
		%$\boldsymbol{E^+} = \{S([a,b],[]),  S([a,a,b,b],[]),$ & & (a)\\
                % 	\qquad \quad $S([a,a,a,b,b,b],[]) \}$ \\
		\multicolumn{2}{l}{$\boldsymbol{E^+} = \{S([a,b],[]), S([a,a,b,b],[]), S([a,a,a,b,b,b],[]) \}$} & (A) \\
		$\boldsymbol{E^-} = \emptyset$ \\
		$\boldsymbol{B} = \{A([a|x], x),  B([b|x],x) \}$ \\
		\midrule
		$\boldsymbol{\mathcal{M}_1} = \{P(x,y) \leftarrow Q(z,u), R(v,w) \}$ & & (B)\\
		$\boldsymbol{\mathcal{M}_2} = \{P \leftarrow Q, R \}$ \\
		\midrule
		\textsc{TOIL-2}$(E^+,E^-,B,\mathcal{M}_1) =$ & $\{P(x,y)\leftarrow Q(x,z),R(z,y) \}$ & (c)\\
		\textsc{TOIL-3}$(E^+,E^-,B,\mathcal{M}_2) =$ & $\{P(x,y)\leftarrow Q(x,z),R(z,y) \}$ & (d) \\
	        \midrule
		$\boldsymbol{\mathcal{M}} = \{P(x,y)\leftarrow Q(x,z),R(z,y)\}$ & & (e)\\
		\textsc{Louise}$(E^+,E^-,B,\mathcal{M}) =$ & $\{ \$1(x,y)\leftarrow S(x,z),B(z,y)$, & (f) \\
							   & $\; S(x,y)\leftarrow A(x,z),\$1(z,y)$, \\ 
							   & $\; S(x,y)\leftarrow A(x,z),B(z,y) \}$ \\
		\bottomrule
	\end{tabular}
\caption{Example of learning with metarules learned by TOIL.}
\label{tab:worked_example}
\end{table}

Table \ref{tab:worked_example} illustrates the use of TOIL to learn metarules
for Louise. In table section (A) the elements of a MIL problem are defined. In
table section (B) a set of matrix metarules $\mathcal{M}_1$ and a set of punch
metarules $\mathcal{M}_2$ are defined, each with a single member. In row (c)
TOIL-2 learns a new fully-connected sort metarule from the elements of the MIL
problem in table section (A) and the matrix metarule in $\mathcal{M}_1$. In row
(d) TOIL-3 learns a new fully-connected sort metarule from the elements of the
MIL problem in table section (A) and the punch metarule in $\mathcal{M}_2$.
Note that both sub-systems of TOIL learn the same fully-connected punch
metarule (the $H^2_2$ \emph{Chain} metarule, listed in Table
\ref{tab:canonical_h22_metarules}).

In rows (e) and (f) Louise is given the elements of the MIL problem in table
section (A) and the metarule learned by TOIL-2 and TOIL-3, and learns the
hypothesis starting at row (f). Note that this is a correct hypothesis
constituting a grammar of the context-free $a^nb^n$ language.

%The example in Table \ref{tab:worked_example} is a real example of learning by
%TOIL and Louise (rather than a simulation ``by hand").

It is interesting to observe that the program learned by Louise includes a
definition of an invented predicate, $\$1$, in row (f). This is despite the fact
that our implementation of TOIL does not perform predicate invention and so has
not learned any metarules that require predicate invention to be learned. In the
MIL problem in Table \ref{tab:worked_example} a single metarule is sufficient to
learn a correct hypothesis and this metarule can be learned without predicate
invention, even though the correct hypothesis starting in (f) cannot, itself, be
learned without predicate invention. This observation suggests that even the
current, limited version of TOIL that cannot perform predicate invention, may be
capable of learning a set of metarules that is sufficient to learn a correct
hypothesis, when given to a system capable of predicate invention, like Louise
(or Metagol). 

The observation about predicate invention in the previous paragraph further
highlights the generality of metarules and suggests the existence of a class of
learning problems that can be solved with a number of metarules much smaller
than the number of clauses in their target theory, a subject for further study.

\end{document}